# Machine Learning-Accelerated Multi-Objective Design of Fractured Geothermal Systems

Guodong Chen[1], Jiu Jimmy Jiao[1,*], Qiqi Liu[2], Zhongzheng Wang[3], Yaochu Jin[2]

[1] Department of Earth Sciences, The University of Hong Kong, Hong Kong, PR China.

[2] College of Engineering, Westlake University, Hangzhou, PR China.

[3] College of Engineering, Peking University, Beijing, PR China.

[*] Corresponding Author: Jiu Jimmy Jiao (jjiao@hku.hk).

## Abstract

Multi-objective optimization has burgeoned as a potent methodology for informed decision-making in enhanced geothermal systems, aiming to concurrently maximize economic yield, ensure enduring geothermal energy provision, and curtail carbon emissions. However, addressing a multitude of design parameters inherent in computationally intensive physics-driven simulations constitutes a formidable impediment for geothermal design optimization, as well as across a broad range of scientific and engineering domains. Here we report an Active Learning enhanced Evolutionary Multi-objective Optimization algorithm, integrated with hydrothermal simulations in fractured media, to enable efficient optimization of fractured geothermal systems using few model evaluations. We introduce probabilistic neural network as classifier to learns to predict the Pareto dominance relationship between candidate samples and reference samples, thereby facilitating the identification of promising but uncertain offspring solutions. We then use active learning strategy to conduct hypervolume based attention subspace search with surrogate model by iteratively infilling informative samples within local promising parameter subspace. We demonstrate its effectiveness by conducting extensive experimental tests of the integrated framework, including multi-objective benchmark functions, a fractured geothermal model and a large-scale enhanced geothermal system. Results demonstrate that the ALEMO approach achieves a remarkable reduction in required simulations, with a speed-up of 1-2 orders of magnitude (10-100 times faster) than traditional evolutionary methods, thereby enabling accelerated decision-making. Our method is poised to advance the state-of-the-art of renewable geothermal energy system and enable widespread application to accelerate the discovery of optimal designs for complex systems.

## Introduction

Climate change has spurred ambitious renewable energy endeavours, dedicated to cultivating a low-carbon, resource-savvy, climate-resilient, and climate-neutral planetary sphere, within which geothermal energy emerges as a notable player in the transition beyond fossil fuels [1,2]. Enhanced geothermal systems (EGS) enable the provision of long-term and sustainable geothermal energy for electricity generation, thus contributing to a fully decarbonized future [3]. The global installed capacity for geothermal power generation currently stands at 16.4 gigawatts electrical (GWe) as of the year-end 2023, with the potential to generate up to 150 GWe of sustainable energy by the year 2050 [4]. However, owning to the high development costs, complex geological uncertainties [5], and geographical limitations, geothermal energy has been undervalued and remains underdeveloped [1,6-8]. Flexible geothermal power has demonstrated its merit in significantly enhancing EGS potential and mitigating electricity supply expenses [2]. The temporal evolution of temperature distributions for such complex systems is described numerically via partial differential equations [9]. Design optimization plays a crucial role in flexible geothermal development processes, as well as in other scientific and industrial fields. Gradient based optimizers have various restrictions when dealing with geothermal design optimization problems, since the underlying partial differential equations make it impossible to calculate the exact gradient of the objective function [10]. Evolutionary computation, such as genetic algorithm and particle swarm optimization, has achieved tremendous progress in optimization, design and modelling across various communities [11], such as hydrogeology [12], computational fluid dynamics [13], automatic control systems [14], and protein engineering [15], due to the global ability to jump out from a local optimum [16]. In renewable energy community, meta-heuristic algorithms also known as evolutionary algorithms have been widely applied to wind turbine air-foil design optimization [17], site selection of solar power plants [18], and geologic $CO_2$ sequestration [19,20]. However, most evolutionary algorithms assume that the evaluation of the objective function comes at no cost [21]. A key issue that impedes the deployment of evolutionary computation on geothermal systems is the prohibitively intensive computational costs required to converge to the global optimum since each calculation of high-fidelity hydrothermal simulation with finite element method is time-consuming.

Machine learning (ML) has recently emerged as a powerful technique for constructing surrogate models, making remarkable advances across various scientific and engineering domains [22-24] and significantly promoting the development of geothermal energy and other energy systems [25,26]. A surrogate model, known as meta-model or proxy, is a computationally cheap and analytically tractable mathematical model constructed using ML models to approximate properties of complex systems [27]. Approaches like artificial neural networks (ANNs), support vector machines, random forests, and Gaussian processes [28,29] can uncover the objective landscapes and identify multi-variable correlations. To overcome the time-consuming nature of high-fidelity simulations, offline ML models are employed as surrogates to predict fluid dynamics represented by partial differential equations [30]. By constructing surrogates, the processes of parameter inversion or design optimization can become computationally efficient, thereby substantially accelerating decision-making [19]. These approaches inevitably suffer from high approximation errors when there are few training samples available, which may mislead the evolutionary search. Deep learning-based surrogate models, encompassing convolutional neural network [31], neural operators [32] and physics-informed neural networks [33], have attained considerable acclaim as an alternative to replace the numerical evolution of partial differential equations [34]. A notable work by Liu, et al. [35] proposed Kolmogorov-Arnold networks, furnishing theoretical assurances for the expressive prowess and neural scaling laws in addressing partial differential equations. Nevertheless, training such globally accurate surrogate for the complex systems necessitates quite a few simulation evaluations, which can prove considerably time-intensive.

In the geothermal energy community, some efforts have been conducted to reduce the number of high-fidelity simulation calculations for efficient optimization purposes [36,37]. Fractures provide preferential flow paths and dominate fluid flow and geothermal energy transport processes because of the higher hydraulic conductivity than surrounding rock [38,39]. Due to the discrepancy between the ground-truth discrete fracture network (DFN) model and the estimated posterior model based on prior information and observed data, it inevitably brings large prediction uncertainty [39]. Besides, the life-cycle management of geothermal reservoirs generally ignores short-term economic benefits, exacerbating great economic uncertainty [40]. The key issues for the development of enhanced geothermal systems are the multidimensional parameters and time-consuming high-fidelity simulations in the optimization and design. Existing evolutionary computation approaches take too long to achieve effective evolution, severely hampering their applications. Prime examples of ML and evolutionary optimization applications towards the challenging issues include random forest as an offline surrogate combined with genetic algorithm for geothermal well placement optimization [41], and deep neural network as an offline surrogate integrated with non-dominated sorting genetic algorithm II (NSGAII) for geothermal well-control optimization [42]. The aforementioned offline surrogate methods trained with one-shot samples replace simulation calculations in the optimization process, which gives limited improvement since offline surrogates are highly dependent on the prediction accuracy. Besides, handling many design parameters remains a critical bottleneck, especially when it comes to computationally intensive physics-driven simulations, which hinders informed real-time decision-making for geothermal design systems and a wide range of complex real-world problems [43,44].

We address this challenge by proposing an Active Learning enhanced Evolutionary Multi-objective Optimization (ALEMO) coupled with hydrothermal simulations in fractured porous media towards efficient fractured geothermal system management (Fig. 1). Since the decision-makers focus more than one target attributes/properties to the EGS (that may have conflicting trends), this study consider multi-objective optimization design to give the trade-offs between multiple objectives. This work expedites the optimization design and automatically provides a set of non-dominated optimal control schemes for EGSs, known as the Pareto front (Fig. 2a). We introduce probabilistic neural network as classifier to pre-screen the promising pareto-optimal solutions. We then use active learning strategy to enhance the local promising (high fitness value) regions of the surrogate model by iteratively infilling samples from the optimum of local surrogate. We validate the performance of ALEMO on 36 benchmark test problems, a fractured geothermal system and a large-scale EGS. We demonstrate the performance of the workflow with optimal non-dominated solutions obtained by only a few hundred numerical simulation calculations, accelerating the optimization process by 1-2 orders of magnitude (10~100-fold faster) than directly using multi-objective heuristic algorithms. Our method promotes the real-time intelligent decision making and development of digital twin system in geothermal field, and enables broad applications in complex design systems like molecular modelling [45,46], electron diffraction [47], battery design optimization [29], and materials design and discovery [48,49].

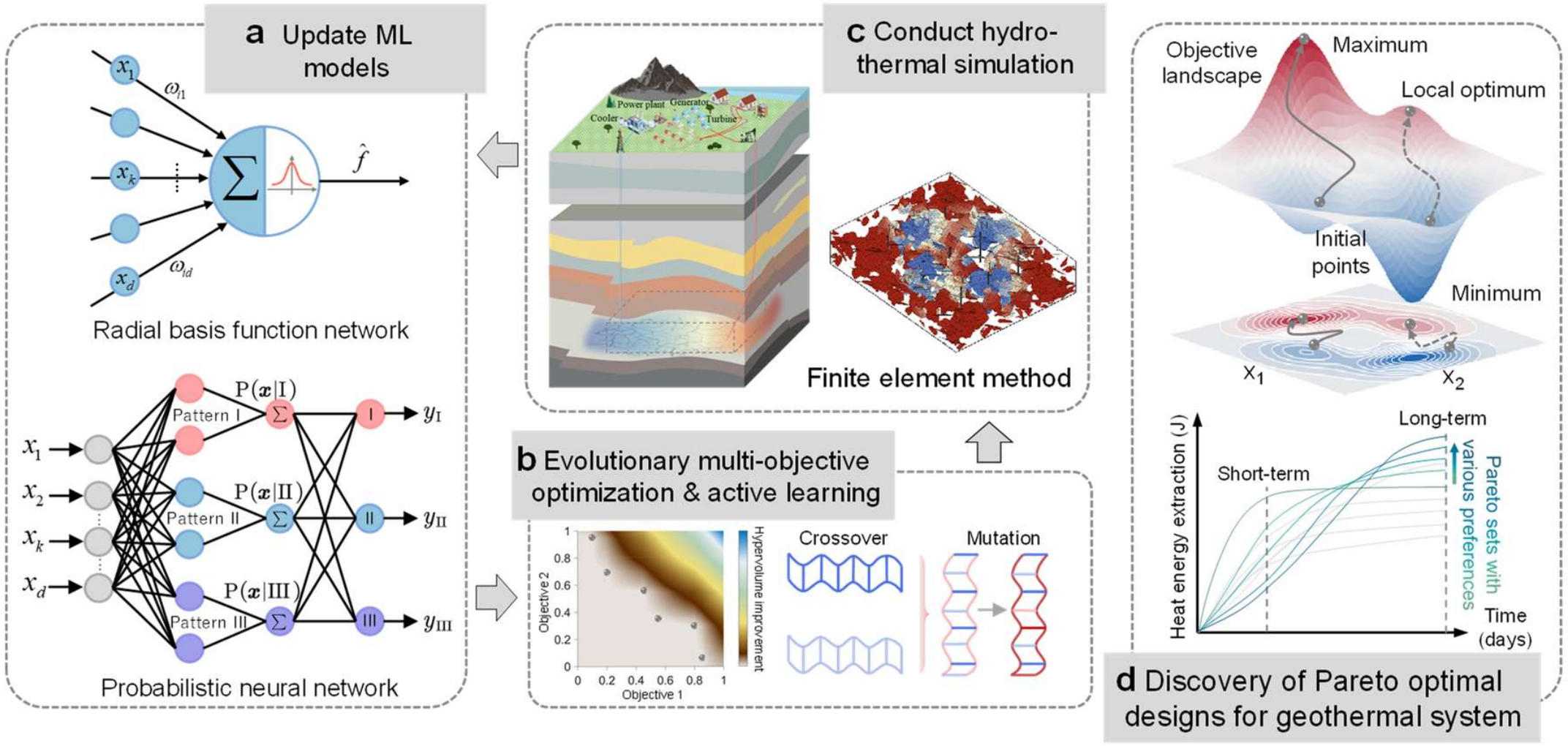

**Fig. 1. Overview of the active learning enhanced evolutionary multi-objective optimization workflow for fractured geothermal systems design. a**, Update ML surrogate models. **b**, Evolutionary multi-objective optimization & active learning. **c**, Hydrothermal simulation using finite element method. **d**, Discovery of Pareto optimal designs for geothermal energy system.

## Multi-Objective Design of Fractured Geothermal Systems

**Multi-objective geothermal problem statement.** The essence of this study is to offer a set of high-quality Pareto solutions for computationally expensive multi-objective numerical functions $\boldsymbol{f}: \mathbb{R}^d \rightarrow \mathbb{R}^m$ with black-box features. An EGS, comprising intricate subsurface and above-ground infrastructure, deploys water injection within high-temperature geothermal reservoirs to rejuvenate natural fractures and extract heated water for electricity generation [38]. The performance evaluation of heat extraction in EGS relies on economic calculations derived from forward hydrothermal simulations. The primary objective in heat extraction optimization is to maximize the net present value (NPV) of the geothermal reservoir throughout the project lifespan, and the design parameters are the water-injection rates and water-production rates (or bottom hole pressure) of wells. Maximizing exclusively the long-term revenue may reduce the short-term performance due to the conflict objectives, whereas solely maximizing the short-term benefits may have a profound adverse influence on the long-term performance. To reduce the risks in the management of geothermal systems, long-term and short-term potential should be simultaneously considered. Thus, given water as the working fluid in this study, the optimization problem can be formulated as:

$$\max_{\boldsymbol{x} \in \mathbb{R}^d} \left[ NPV_l(\boldsymbol{x}, \boldsymbol{z}_l), NPV_s(\boldsymbol{x}, \boldsymbol{z}_s) \right]$$

$$NPV(\boldsymbol{x},\boldsymbol{z}) = \sum_{i=1}^{N_t} \frac{\Delta t_i}{(1+\gamma)^{\frac{\Delta t}{365}}} [TEPR_i \times r_e - WIR_i \times r_i - WPR_i \times r_p]$$

$$s.t. \begin{cases} \boldsymbol{lb} \leqslant \boldsymbol{x} \leqslant \boldsymbol{ub} \\ \boldsymbol{c}(\boldsymbol{x}) \leqslant 0 \end{cases}$$

with $\boldsymbol{x}$ well injection-production variables, $\boldsymbol{z}$ state variables such as temperature and pressure derived from hydrothermal simulation, subscript *l* and *s* long term and short term, $TEPR_i$, $WIR_i$, and $WPR_i$ thermal energy production rate, water injection rate and production rate at the i^th^ time step for the EGS, respectively, $\Delta t_i$ the i^th^ timestep length, $\gamma$ discount rate, $N_t$ the total number of life-cycle timesteps, $r_e$, $r_i$ and $r_p$ thermal energy price per watt-hour for electricity generation [50], water injection and production cost, respectively, $\boldsymbol{lb}$ and $\boldsymbol{ub}$ lower bounds and upper bounds of decision variables, $\boldsymbol{c}(\boldsymbol{x})$ constraints of decision variables, such as physical and operational constraints.

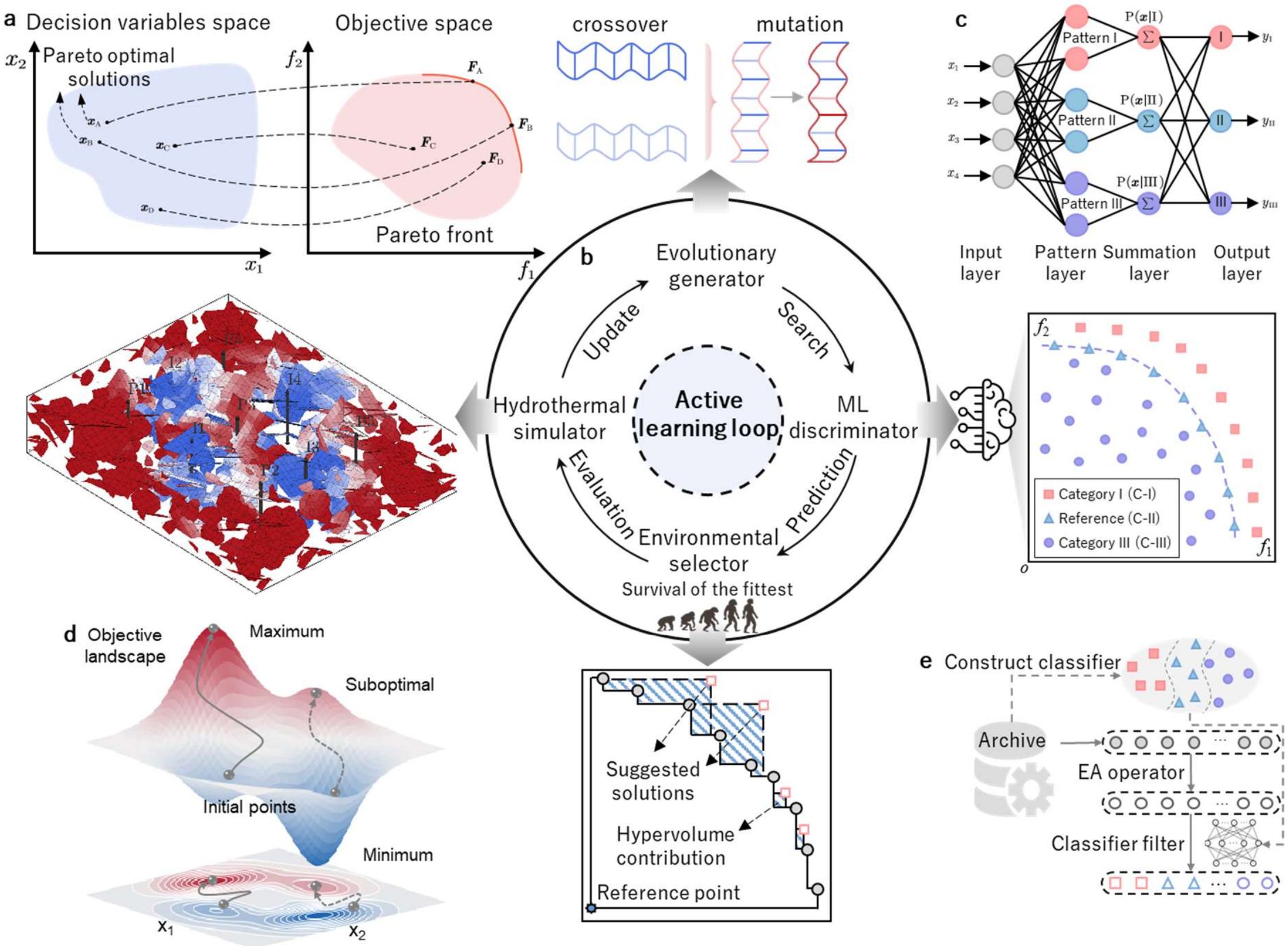

**Fig. 2. Architecture of the active learning enhanced evolutionary multi-objective optimization for fractured geothermal systems design. a**, Schematic illustration of multi-objective problems and related concepts. **b**, Closed-loop active learning for ALEMO consists of four components, evolutionary generator, ML discriminator, environmental selector, and hydrothermal simulator. Evolutionary generator generates offspring using mutation and crossover operator; ML discriminator identifies elite solutions from offspring; Environmental selector chooses the most promising solution based on the prediction of surrogate model; Hydrothermal simulator calculates the true objective value. **c**, Schematic

diagram of PNN, featuring a pattern layer and a summation layer, effectively assessing pattern probability and discerning intricate patterns with promising precision. **d**, Active learning enhanced surrogate-directed evolutionary optimization process. **e**, Framework of classifier-assisted evolutionary search.

**ALEMO approach overview.** Schematic of the proposed active learning enhanced evolutionary multi-objective optimization workflow for fractured geothermal systems design is shown in Fig. 2b. Our approach ALEMO consists of two ingredients: discriminator-assisted evolutionary exploration, and hypervolume-based attention subspace search. In this study, we focus on not only efficiently, but confidently, identifying the Pareto optimal solutions. To accomplish this goal, we employ an active learning strategy that enables iteratively interacting with and learning from the high-fidelity hydrothermal simulation to refine ML models. Active learning strategically leverages uncertainty information and the identified regions within high-fitness value to explore the parameter space and reduce uncertainty with as few steps as possible [51], thus ensuring the balance between exploration and exploitation throughout the optimization process [52,53]. See Supplementary Fig. S2 for comparison of offline optimization with one-shot sampling and active learning enhanced online optimization. The new training data is sequentially and strategically queried for efficiently discovering Pareto optimal solutions or close to Pareto optimal solutions without significantly increasing the size of training dataset [46,54]. The loop ceases when the additional infilling samples no longer contributes to the information gain, i.e., the Pareto optimal solutions remain unchanged. At the end of the optimization process, a set of optimal solutions is supplied, known as the Pareto front (Fig. 2a). These optimal or non-dominated solutions cannot be further improved in terms of one objective, such as maximizing long-term NPV, without compromising the corresponding target value.

**Discriminator-assisted evolutionary exploration.** The main goal of discriminator-assisted evolutionary exploration is to identify the most informative infill samples and guide the multi-objective search process assisted by the discriminator. We construct a discriminator using PNN (Fig. 2c) with the classified solutions that learns to predict the dominant relationship between candidate samples and reference samples. The classification criterion holds utmost significance within classifier-assisted evolutionary exploration, as the infill samples filtered by classifier heavily reliant on the classification criterion. The classification criterion in this work for training the discriminator is illustrated in ML discriminator of Fig. 2b. The solutions in the second non-dominated Pareto front are set as the reference solutions and are classified into category II. For the solutions that are dominated by the reference solutions are classified into category III, while for the solutions dominate the reference solutions are classified into category I (Fig. 2b and Supplementary Fig. S1). Detailed discriminator-assisted evolutionary exploration (Fig. 2e) comprises the following parts:

(1) Initialization. We employ Latin hypercubic sampling as design of experiment approach to sample uniformly from the design space and conduct high-fidelity hydrothermal simulations as the initial samples in the archive $\boldsymbol{A}=\{\boldsymbol{x}_i,\boldsymbol{y}_i\}$ (Archive *A* is for collecting the evaluated samples and constructing surrogate models).

(2) Discriminator management. We classify the solutions into three categories based on pre-defined classification criterion in Fig. 2b, and train or update the weights of the discriminator with the training data.

(3) Offspring generation. We conduct non-dominated sorting for the solutions in the archive $\boldsymbol{A}$, and select top *NP* solutions as the population $\boldsymbol{P} \leftarrow \{\boldsymbol{x}_1, ..., \boldsymbol{x}_{NP}\}$ based on non-dominated sorting and crowding distance metrics [11]. Solutions in the second level are set as reference solutions and labelled C-II. Solutions dominate reference solutions are labelled C-I, while solutions dominated by reference solutions are labelled C-III. After that, we construct discriminator to classify the offspring and distinguish promising from poor samples. We then develop rank-based learning operator to generate more promising and informative candidate solutions:

$$\boldsymbol{v}_i = \boldsymbol{x}_{r\mathrm{I}} + Mu \times (\boldsymbol{x}_{r\mathrm{I}} - \boldsymbol{x}_{r\mathrm{II}})$$

where $\boldsymbol{x}_{r\mathrm{I}}$ and $\boldsymbol{x}_{r\mathrm{II}}$ are the solutions randomly selected from the C-I and C-II level, and *Mu* is the mutation factor. Polynomial mutation and crossover operation are then performed to generate offspring $\boldsymbol{u}$.

(4) Selection and simulation evaluation. We employ the discriminator to rank the offspring candidates. The uncertainty of offspring solutions classified to the first level is represented by the minimum Euclidean distance between the offspring solutions and the sampled solutions in the archive:

$$g(\boldsymbol{u}_i) = \min_{\boldsymbol{x} \in \boldsymbol{x}_{\mathrm{I}}} \{dis(\boldsymbol{u}_i, \boldsymbol{x})\}$$

with $dis(\boldsymbol{u}_i, \boldsymbol{x})$ the Euclidean distance in decision space between the offspring $\boldsymbol{u}_i$ and evaluated sample points $\boldsymbol{x}$ in $\boldsymbol{A}$. Larger values of g(x) indicate that the solution x in the decision space is sparser, and the greater the uncertainty. The most uncertain offspring is selected for exploration:

$$\hat{\boldsymbol{x}}_c = arg \max_{\boldsymbol{u}_i \in \boldsymbol{P}} g(\boldsymbol{u}_i)$$

with $\hat{\boldsymbol{x}}_c$ the selected solution to be evaluated. After that, high-fidelity hydrothermal simulation is conducted as the active learning loop. New data is actively queried for efficiently improving the approximation quality of the surrogate and accelerating the optimization process.

**Hypervolume based attention subspace search.** In this stage, we identify a small attention subspace and construct surrogates for the objectives in the attention subspace. To efficiently exploit and attention to the current promising subspace, we determine the boundary of the attention subspace by locating the subspace of the $\tau$ promising solutions $\boldsymbol{A}_\tau$ selected by non-dominated sorting and crowding distance metrics:

$$\begin{cases} \boldsymbol{lb}_s = \min(\boldsymbol{x}_1, \boldsymbol{x}_2, ..., \boldsymbol{x}_\tau) \\ \boldsymbol{ub}_s = \max(\boldsymbol{x}_1, \boldsymbol{x}_2, ..., \boldsymbol{x}_\tau) \end{cases}$$

with $\boldsymbol{lb}_s$, $\boldsymbol{ub}_s$ the lower and upper bounds of the attention subspace. The identification of attention subspace offers a straightforward way to avoid exhaustive searches in the original vast parameter spaces. The surrogate $\hat{\boldsymbol{f}}(\boldsymbol{x})$ for the attention subspace is constructed for the objectives via weighted sum of kernel functions $\sum_{i=1}^{n}\lambda_i\varphi(\|\boldsymbol{x}-\boldsymbol{c}_i\|)$ with Gaussian radial-basis function kernels $\varphi(r)=\exp\left(-\frac{r^2}{\sigma^2}\right)$. The weight coefficient can be determined as $\boldsymbol{\lambda}=\boldsymbol{\Phi}^{-1}\boldsymbol{y}$, where $\boldsymbol{\Phi}=[\varphi(\|\boldsymbol{x}_i-\boldsymbol{x}_j\|)]_{\tau\times\tau}$ is the kernel matrix. We then employ evolutionary mutation, polynomial mutation and crossover operator to generate offspring and update populations via non-dominated sorting approach [55], until the approximated Pareto front of the attention subspace is identified. The final infill solution is selected via the criterion:

$$\hat{\boldsymbol{x}}_h = arg\max_{\hat{\boldsymbol{x}}_i\in\boldsymbol{A}_\tau}[HV(\boldsymbol{x}_{\boldsymbol{A}_\tau}\cup\hat{\boldsymbol{x}}_i)-HV(\boldsymbol{x}_{\boldsymbol{A}_\tau})]$$

with $\hat{\boldsymbol{x}}_i$ the $i^{th}$ approximated Pareto solution for the attention subspace, and $HV$ the hypervolume improvement value to measure the dominated volume of the solutions in the objective space. The most promising solution $\hat{\boldsymbol{x}}_h$ is actively queried using high-fidelity hydrothermal simulation by maximizing the hypervolume improvement value. Notably, by actively querying the candidate solutions wisely, theoretical guarantees of identifying the set of Pareto optimal control schemes can be obtained consistently.

## Results

We first conduct experiments on Deb-Thiele-Laummans-Zitzler (DTLZ) [56] and Zitzler-Deb-Thiele (ZDT) [57] multi-objective benchmark functions that are commonly used in evolutionary computation community, including functions with shifted, non-separable, concave, disconnected and multi-modal characteristics. To examine the performance of ALEMO in the benchmark problems, various methods were tested, such as canonical evolutionary multi-objective methods NSGAII and MOEAD, as well as state-of-the-art machine learning assisted methods, such as CPS-MOEA, MCEA/D, and CSEA. In the experiments, all methods were performed 300 function evaluations in each trial, for a total of 20 independent trials to analyse the statistical results. We conducted experiments on 36 multi-objective benchmark functions and two fractured geothermal systems to demonstrate the effectiveness of the proposed approach. In the first experiment, the dimension of the benchmark functions ranges from 10 to 30, and the number of objectives is two and three.

**Experiments on multi-objective benchmark functions.** We examine ALEMO framework through an exploration of 36 multi-objective benchmark functions that are widely employed in the realm of multi-objective optimization (See Supplementary Table S-I and S-II for detailed function expressions and characteristics of the DTLZ and ZDT multi-objective benchmark problems). These functions span a gamut of characteristics, ranging from unimodal and multimodal to separable and non-separable, linear and concave, and even encompassing instances that exhibit many-to-one as well as disconnected traits. Further elaboration on these benchmark functions can be found in Supplementary Table 1. As we mentioned, a key challenge of canonical multi-objective evolutionary methods is the vast parameter space and multiple objectives involved in the optimization and iteratively design. Averaged convergence curves and comparison results of ALEMO and various methods on 30-D benchmark problems are derived via 20 independent trials with random initializations (See Supplementary Fig S3-4 for detailed optimization and comparison results of ALEMO and baseline methods on 10-D and 20-D multi-objective benchmark problems).

The heuristic approaches contain stochastic properties, and their main theoretical guarantee is how many computational resources are required to attain optimal Pareto solutions or near-optimal non-dominated solutions or how the approach converges with given computational resources. Canonical methods NSGAII and MOEAD are slow to converge and do not work well compared to other data-driven surrogate-based methods, because surrogate-based methods leverage the prediction capability of ML for approximation of objective landscapes to guide evolutionary search. From the convergence results in Fig.3, ALEMO attains the best performance in most benchmark problems. Since the benchmark functions contain various characteristics, it is nontrivial to achieve best performance in all the benchmark problems. One critical challenge of the design optimization problems is the multi-dimensional parameter space. Similar ranking results are attained with dimension varying from 10 to 30 in Fig. 3m-o. Empirical ranking of different methods in 10-D, 20-D, and 30-D benchmark problems shows the scalable performance of ALEMO on more challenging high-dimensional problems. See Supplementary Fig. S3-4 for averaged convergence performance of data-driven evolutionary computation approaches on 10D and 20D benchmark problems with 20 independent trials.

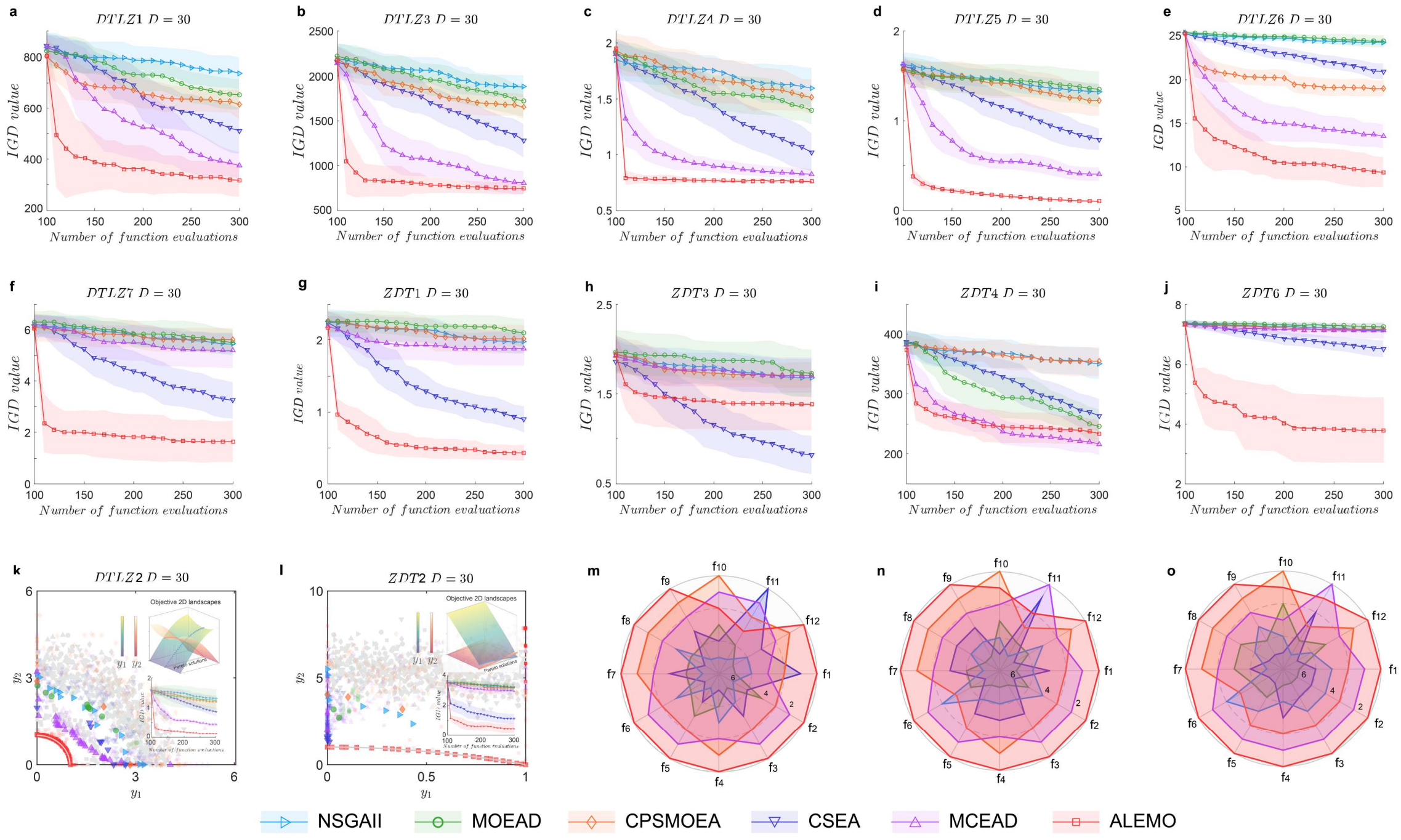

**Fig. 3. Accelerated multi-objective optimization in benchmark problems. a-j**, Averaged convergence performance of data-driven evolutionary computation approaches with 20 independent runs presented as mean±s.e.m.. **k-l**, Objective landscapes, averaged convergence performance and final non-dominated solution distributions of data-driven evolutionary computation approaches in DTLZ2 ($f_2$) and ZDT2 ($f_9$) benchmark problems. **m-o**, Empirical ranking of different methods in 10-D, 20-D, and 30-D benchmark problems.

From Fig. 3k, l, ALEMO converges efficiently and locates the Pareto front of DTLZ2 and ZDT2 benchmark problems with promising diversity of solutions in given limited number of function evaluations, whereas canonical evolutionary computation methods and other data-driven surrogate-based methods haven't converged nearby Pareto front yet. The efficient convergence of ALEMO can be attributed to the hypervolume based attention subspace search, while the promising diversity for the final non-dominated solutions of ALEMO can be attributed to the discriminator-assisted evolutionary exploration. The discriminator enables us to determine which solution can be confidently discard, and which solution can maximally improve the current Pareto front. After attention to the promising subspace, the optimizer locates new queried solutions by employing hypervolume indicator as the infill criterion. The cooperation of hypervolume based attention subspace search and discriminator-assisted evolutionary exploration enables efficiency and effectiveness of the proposed method. We take advantage of active learning for model management to strategically use uncertainty information and identify objective regions within high-fitness value to explore the parameter space and reduce uncertainty with as few steps as possible, thus ensuring the balance between exploration and exploitation throughout the optimization process.

**Multi-objective design for fractured geothermal energy system.** Geothermal energy system design presents an essential yet complex engineering challenge in attaining sustainable and decarbonized electricity networks with negligible environmental impact. See Supplementary Fig. S5 for schematic diagram of the geothermal power generation systems. Geothermal reservoir simulation and design optimization determines injection and production flow rates and bottomhole pressure (BHP) at constant interval to ensure compliance with the physical constraints and maximize geothermal energy development. We consider multi-objective techno-economic optimization for the geothermal reservoir to make trade-off between short-term and long-term economic profits. Based on the prior information of the geothermal energy system, we construct a multi-physics numerical simulation model by integrating fluid flow and heat transfer process in fractured porous media using the numerical simulator COMSOL. The detailed fractured geothermal energy system and corresponding DFN information and well-placement distribution are shown in Fig. 4a-c. The involved mass balance and heat transfer governing equations are calculated by finite element method.

We consider a simplified model configuration as a 2 km × 2 km geothermal reservoir consisting of 9 wells (5 producers and 4 injectors) at the depth of 2500m with an initial temperature of 473.15K. Reservoir cooling is predominantly propelled by fluid advection linked to the injection of cold water, consequently relying primarily on the flow rate. Therefore, the flow rate exerts significant influence in delineating the alteration in reservoir temperature. For the geothermal reservoir, producers are set to constant BHP as initial reservoir pressure 30 MPa. The parameters that need be manipulated are the injection rate of the injectors at each time step of 600 days for a period up to 12000 days, a total of 80 variables. Operation on injection rates of injectors enables heat extraction from the producers by controlling and changing fluid flow constantly. Considering only maximizes the long-term economic profits of the geothermal energy system may cause high risks due to the uncertainty of the constructed simulation model, we optimize long-term and short-term economic profits of the geothermal reservoir simultaneously. See Table S-III for detailed parameter configurations of the geothermal reservoir.

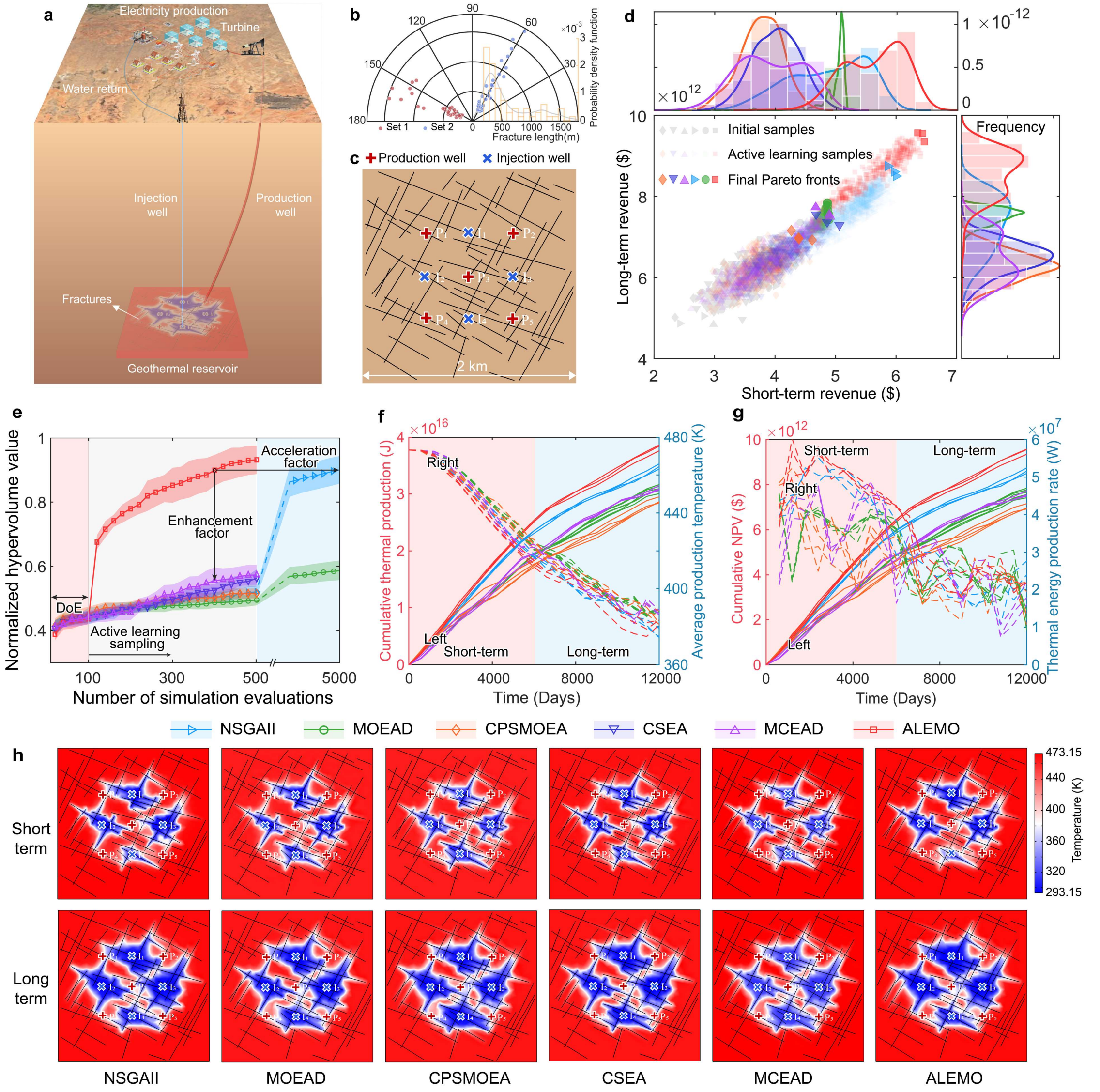

**Fig. 4. Accelerated multi-objective geothermal energy system design.** **a**, Enhanced geothermal system components for power generation process. **b**, Statistical information of the DFN model, including fracture geometry, length and orientation. **c**, DFN model description and well-placement distribution. **d**, The empirical Pareto fronts for the various multi-objective methods for the trade-off between short-term and long-term economic profits of the geothermal energy system. **e**, Hypervolume convergence curves of the six multi-objective methods on the geothermal energy system design problem. **f**, Simulation performances of cumulative thermal production and average production temperature for the optimized well-control schemes provided by various multi-objective methods. **g**, Simulation performances of cumulative NPV and thermal energy production rate for the optimized well-control schemes provided by various multi-objective methods. **h**, Short-term and long-term temperature distributions for the optimized well-control schemes with highest long-term revenue provided by various multi-objective methods.

We give canonical multi-objective methods NSGAII and MOEAD with ten times more computational resources than data-driven surrogate-based methods to understand to what extent of the proposed method accelerates the optimization process. In the case of machine learning guided evolutionary methods, the total number of simulation evaluations is set at 500, whereas for the classical NSGAII and MOEAD, it is set at 5000. Fig. 4d illustrates the empirical Pareto fronts for the various multi-objective methods for the trade-off between short-term and long-term economic profits of the geothermal energy system. The proposed ALEMO obtains better Pareto front with higher long-term and short-term economic profits than other methods (Fig. 4d). Hypervolume convergence curves of the six multi-objective methods on the geothermal energy system design problem (Fig. 4e) present the quantitative optimization results, showing superior performance of the proposed ALEMO. ALEMO accelerates the optimization computation approximately 10 times more efficiently than NSGAII and 50 times more efficiently than MOEAD.

Fig. 4f demonstrates simulation performances of cumulative thermal production and average production temperature for the optimized well-control schemes provided by various multi-objective methods. Simulation results demonstrate the continuous decrease in average thermal energy production temperature and continuous increase in cumulative thermal energy production by controlling the injection rate of four injectors. Simulation performances of cumulative NPV and thermal energy production rate for the optimized well-control schemes provided by various multi-objective methods are presented in Fig. 4g. Cumulative NPV shows similar trend with cumulative thermal production since the costs of water injection and production are negligible for this model. Thermal energy production rate of ALEMO remains relatively higher than other methods throughout the project period. The control schemes provided by the proposed ALEMO can achieve higher cumulative thermal production and NPV, enabling not only short-term but also long-term economic profits. Fig. 4h demonstrates the detailed short-term and long-term temperature distributions for the optimized well-control schemes with highest long-term revenue provided by various multi-objective methods. See Supplementary Fig. S6 for results of ALEMO and evolution of thermal dynamics of the fractured geothermal reservoir.

**Field-scale EGS multi-objective design.** After demonstrating that our workflow is capable of accelerating the discovery of optimal multi-objective geothermal energy system design effectively, we further conduct techno-economic multi-objective design on a more challenging field-scale EGS. EGS technologies primarily focus on subsurface formations with substantial geothermal potential yet low hydraulic permeability in the host rock [2]. Fig. 5a presents the schematic diagram of the EGS components for power generation process. Three-dimensional DFN is delineated under the constraints of the collected prior information of geothermal field, encompassing 1258 fractures (See Fig. 5b for fracture orientation and azimuth distribution of the fractured geothermal energy system).

The embedded discrete fracture network method provides efficient but accurate way to account for such field-scale fractures with coarse and regular matrix cell. Besides, optimization experiments are inherently time-consuming. Each hydrothermal simulation in fractured porous medium requires substantial computational resources, and optimizations typically involve hundreds of simulation evaluations. Conducting multiple trials and employing various methods is necessary to obtain statistically meaningful results and ensure equitable comparisons. Thus, the geothermal energy system is discretized numerically utilizing the embedded discrete fracture network method, facilitating the examination of fluid flow and heat transfer within the DFN and host rocks through the application of the finite volume method using MRST software [58]. Our field-scale geothermal model comprises 5 injection wells and 3 production wells, encompassing a reservoir domain 1000 m × 500 m × 100 m at a depth of 3.5 km (Fig. 5c). The initial temperature of the EGS is $473.15\ \mathrm{K}$, while the temperature of the injected cold water stands at $323.15\ \mathrm{K}$. This geothermal system operates the injection bottomhole pressure of injection wells and production flow rates of production wells at constant intervals within the physical constraints to enhance the exploitation of geothermal energy. The parameters requiring optimization encompass the injection BHPs of the injection wells and production rates of the production wells at each time step spanning 300 days for a period up to 6000 days, a total of 160 variables. The simulation model is designed through multi-objective techno-economic optimization for the EGS, ensuring a delicate balance between short-term and long-term economic profits.

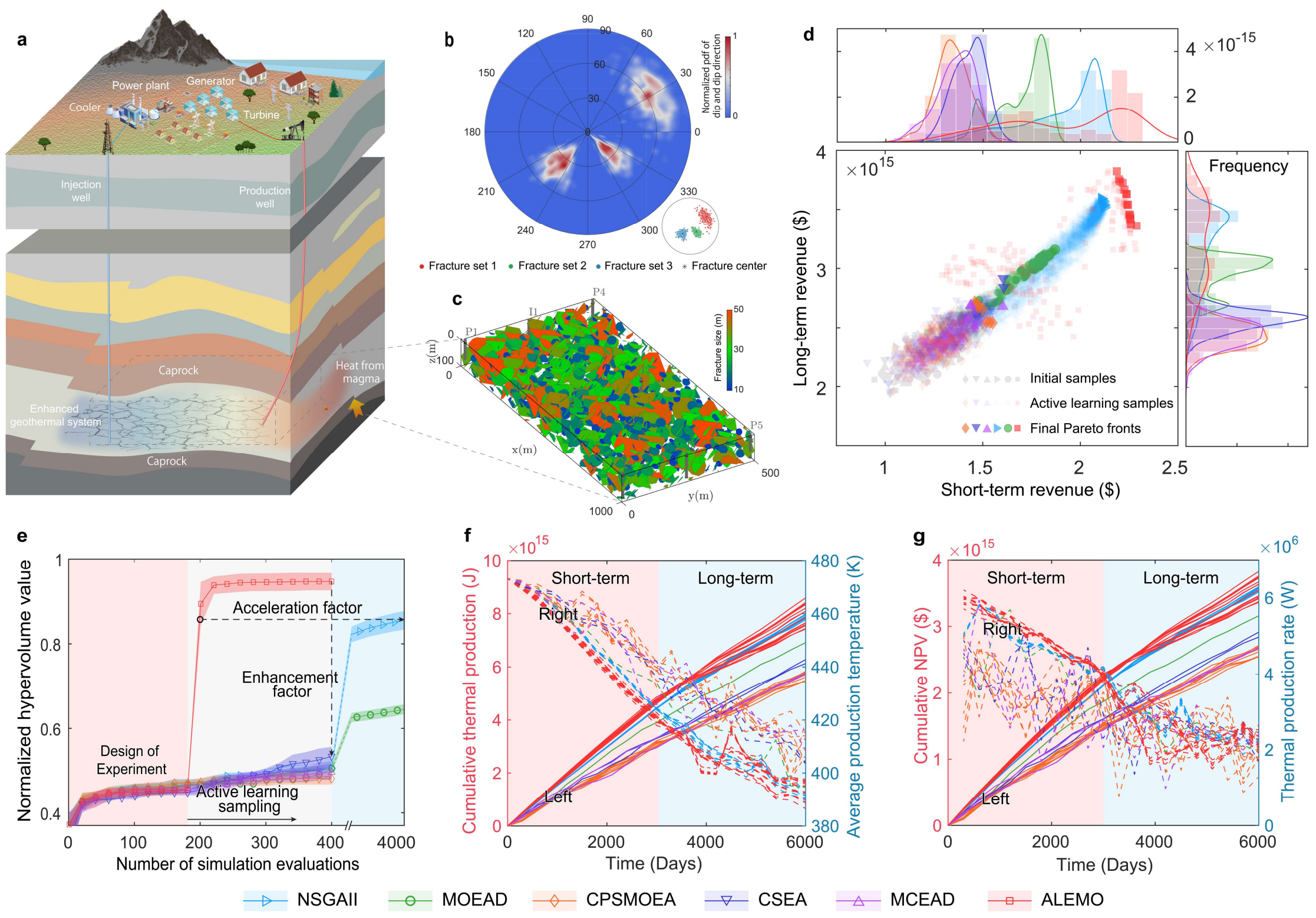

**Fig. 5. Accelerated discovery of optimal field-scale multi-objective fractured geothermal energy system design. a**, Schematic diagram of EGS components for power generation. **b**, Fracture orientation and azimuth distribution for the fractured geothermal energy system. **c**, Three-dimensional field-scale discrete fracture network and well placement distributions. **d**, Empirical Pareto fronts for the various multi-objective methods for the trade-off between short-term and long-term economic profits of the field-scale fractured geothermal energy system. **e**, Hypervolume convergence curves of the various multi-objective methods on the field-scale geothermal energy system design problem. **f**, Simulation performances of cumulative thermal production and average production temperature for the Pareto

well-control schemes provided by various multi-objective methods. **g**, Simulation performances of cumulative NPV and thermal energy production rate for the Pareto well-control schemes provided by various multi-objective methods.

In the case of machine learning-directed evolutionary approaches, the total count of simulation evaluations is stipulated as 400, whereas for classical NSGAII and MOEAD, we conduct ten-fold model evaluations to validate the acceleration effect of the proposed method. Fig. 5d-e illustrates the empirical Pareto fronts for the various multi-objective methods and corresponding averaged hypervolume convergence curves under 5 independent trials for the trade-off between short-term and long-term economic profits of the field-scale fractured geothermal energy system. We compare our method with other classical and state-of-the-art multi-objective optimization algorithms in the literature, and find that our method outperforms these algorithms. Our method greatly accelerates the optimization speed and reduces the required model evaluations by dozens of times compared to baseline optimization approaches (Fig. 5e). The simulation performances of cumulative thermal production, average production temperature, cumulative NPV and thermal energy production rate for the Pareto well-control schemes provided by our method with other baseline optimization approaches (Fig. 5f-g) show the final Pareto solutions of our method achieve high cumulative thermal energy production, cumulative NPV and thermal energy production rate throughout the project period. The average production temperatures of the fluids for the Pareto well-control schemes of ALEMO keep a low standard throughout the project period, and decrease from 473.15 K (day 0) to between 380 K and 400 K after 6000 days (Fig. 5f). The empirical Pareto fronts of our method and other baseline approaches (Fig. 5d) demonstrate that the accelerated efficiency of our method does not come at the expense of accuracy. See Supplementary Fig. S7 for simulation results of ALEMO and compared baseline methods.

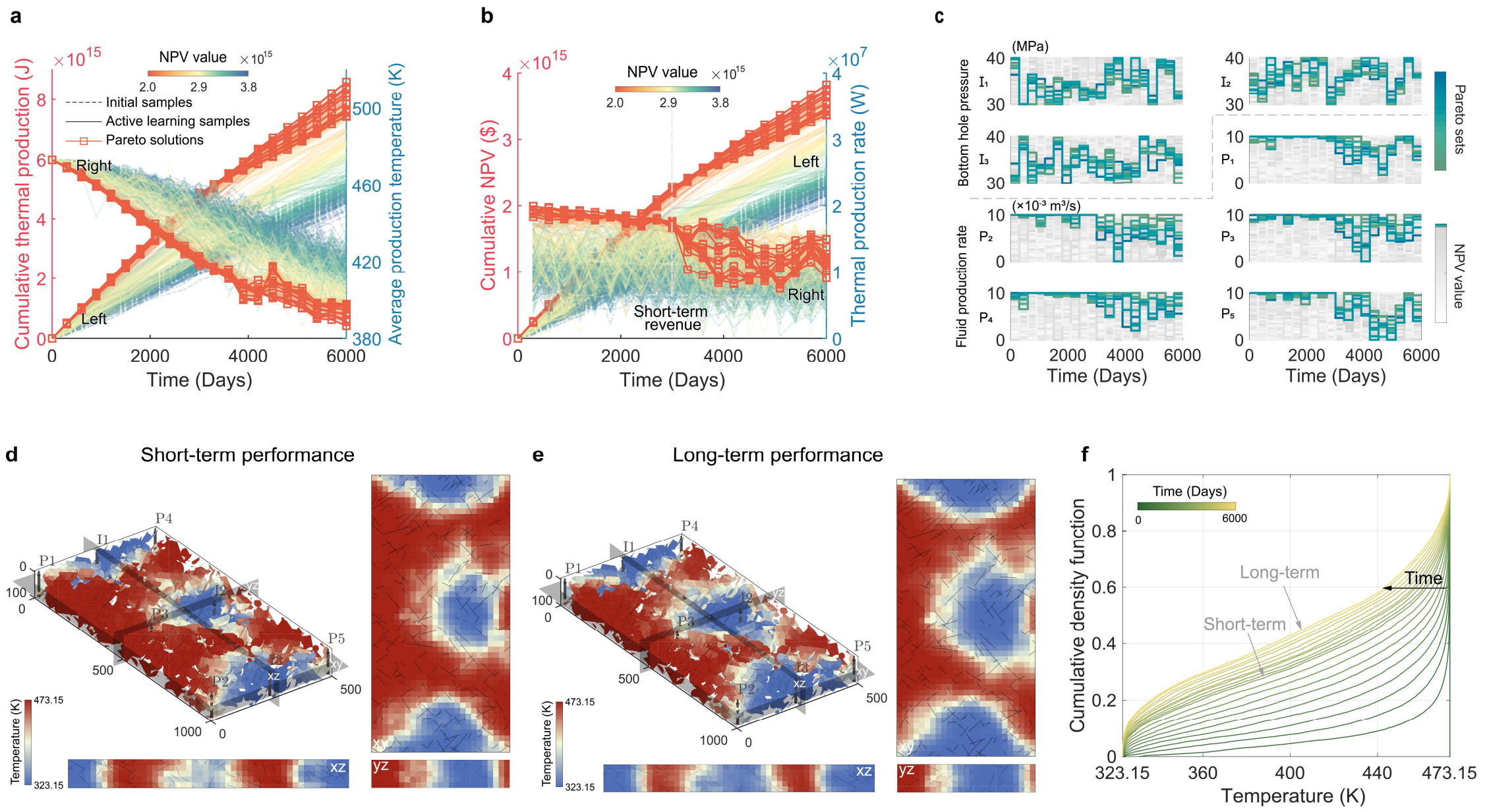

**Fig. 6. Pareto optimal design of ALEMO and evolution of thermal dynamics of the field-scale enhanced geothermal energy system. a**, Simulation results of the cumulative thermal energy production and average production temperature of the enhanced geothermal energy system for the

sampling history of ALEMO. **b**, Simulation results of the cumulative net present value and thermal energy production rate for the sampling history of ALEMO. **c**, Well control trajectories and corresponding net present value of the sampling history of ALEMO. **d**, Visualization of short-term temperature distribution for the simulation model evaluations of ALEMO with the highest long-term revenue. **e**, Visualization of long-term temperature distribution for the simulation model evaluations of ALEMO with the highest long-term revenue. **f**, Temperature evolution process for the simulation model evaluations of ALEMO with the highest long-term revenue during the project period.

We validate the performance of ALEMO with simulation results and evolution of thermal dynamics of the field-scale EGS (Fig. 6). From Fig. 6a-b, simulation results of the cumulative thermal energy production, average production temperature cumulative net present value and thermal energy production rate of the enhanced geothermal energy system for the entire sampling history of ALEMO are presented. The objective values of the initial DoE samples (dashed line) are relatively low, while the objective values of the machine learning-directed samples are significantly improved. The new samples are actively queried for efficiently improving the approximation quality of the surrogate model without significantly increasing the size of training samples. The final Pareto solutions show great diversity and significant improvement, providing versatile choice for decision-makers. Well control trajectories of the sampling history of ALEMO (Fig. 6c) show the parameter space distribution of the final Pareto solutions with different preferences. We visualize the short-term and long-term temperature distribution for the simulation model evaluations of ALEMO with the highest long-term revenue (Fig. 6d-e). In this reference simulation, the initial average production temperature is 473.15 K, and declines over time to between 380 K and 390 K after 6000 days. The electricity generation decreases from 20 MW (day 0) to about 10 MW (day 6000). From Fig 6f, the temperature decreases shapely at the early period and decrease trend becomes gentle in the later stage. The enhancement factor and acceleration factor primarily arise from the support provided by the discriminator and the surrogate model, which enable the selection of more promising and informative offspring solutions generated by evolutionary algorithms for real function evaluations. This strategically reduces unnecessary evaluations in low-fidelity areas. Additionally, the discriminator-assisted evolutionary exploration identifies the most informative infill samples, thereby enhancing the overall multi-objective optimization process and preventing falling into local optima. This result empowers the selection of designs contingent upon predilection of decision-makers, and enables the scalability and prospective applications in intricate system design conundrums.

This research is applicable for the optimal design of geothermal energy systems, encompassing practical application aspects such as well injection and production rates, well placement, and hydraulic fracturing design parameters, thereby directly impacting the design and operation of geothermal systems and enhancing the performance of geothermal power generation. Besides, this cutting-edge research not only advances the state-of-the-art in renewable geothermal energy systems but also exemplifies the importance of interdisciplinary collaboration in bridging gaps between research, development, and practical implementation. By enhancing our understanding of complex systems, our method paves the way for accelerated decision-making, with significant impacts on sustainable energy development and a broader readership in the fields of engineering, environmental science, and energy policy. Ultimately, this work positions itself as a valuable contribution to the nexus of artificial intelligence and real-world application, facilitating the discovery of optimal designs that support a sustainable future.

## Discussion

We have reported the active learning enhanced evolutionary multi-objective optimization algorithm, integrated with hydrothermal simulations in fractured media, to expedite the design of geothermal systems in multiple orders of magnitude. Our workflow comprises two primary components: (1) discriminator-assisted evolutionary exploration, wherein the most informative infill samples are identified under the guidance of the discriminator, thereby expediting the multi-objective optimization process; and (2) hypervolume based attention subspace search, enabling the identification of a focused attention subspace and construction of surrogates for the objectives. We incorporate the active learning strategy to substantially improve the performance of machine learning model by iteratively updating models and infilling new informative samples. The new training data is sequentially and strategically queried for efficiently discovering Pareto optimal solutions or close to Pareto optimal solutions with fewest number of samples. Results on 36 multi-objective benchmark test problems, and two enhanced geothermal systems demonstrated our approach accelerates 1-2 orders of magnitude simulation evaluations.

Although this work is performed in the context of fractured geothermal system design applications, ALEMO is not limited to geothermal heat extraction optimization but has great potential for widespread applications in the design and optimization of complex systems like oil reservoir production optimization, aerodynamic shape optimization, and materials design and discovery. The modular framework of ALEMO enables customizable adjustments in various systems and allows users to integrate other technologies, such as parallel computation and uncertainty analysis, to enhance the comprehensive understanding of the system. More generally, this work paves the way to design optimization and accelerated decision-making in complex systems like fractured geothermal systems, especially for the problems with vast parameter spaces. We expect that in the foreseeable future, by drawing on the success of large language models, large geothermal domain model could be incorporated to enable digital representations of complex physical geothermal systems and provide real-time geological parameter estimation and well-control. Considering geo-mechanical effects can significantly impact simulation results, we will incorporate geo-mechanical effects in our future research by employing reinforcement learning for decision-making, alongside implementing thermal destressing strategies to reduce the risk of large earthquakes propagating through critical states.

## Methods

**Mass conservation equation.** The mass conservation equation of subsurface groundwater flow in porous and DFN media for geothermal reservoir is given as

$$\frac{\partial}{\partial t}(\rho_w \phi) + \nabla(\rho_w u_m) = \Psi_{mf} + Q_m$$

with $t$ time, $\rho_w$ water density, $\phi$ porosity, $\Psi_{mf}$ the flux transfer function between fractures and matrix, $Q_m$ mass source/sink term, and $u$ the velocity of fluid defined as

$$u = -\frac{k}{\mu}(\nabla p + \rho_w g \nabla D)$$

with $k$ permeability (fracture permeability is defined by cubic law $k_f = b^2/12$, $b$ is fracture aperture), $\mu$ the water viscosity, $p$ pressure, $g$ gravitational acceleration, and $\nabla D$ a unit vector along the direction of gravity. The flux transfer function $\Psi_{mf}$ writes

$$\Psi_{mf} = \nabla \cdot \left(\rho_w \cdot \left(-\frac{k}{\mu} \nabla P_{boundary}\right)\right)$$

with $\nabla P_{boundary}$ the pressure gradient in the tangential plane of a fracture.

**Energy conservation equation.** Heat transport in the matrix of EGS is described by the energy conservation equation as

$$(\rho C)_{eff} \frac{\partial T}{\partial t} + \rho_w C_w u_m \cdot \nabla T - \nabla \cdot (\lambda_{eff} \nabla T) = E_{mf} + Q_{hm}$$

with $C$ the specific heat capacity, $(\rho C)_{eff} = \phi \rho_w C_w + (1-\phi)\rho_s C_s$ the effective heat capacity, $\lambda_{eff} = \phi \lambda_w + (1-\phi)\lambda_s$ the effective thermal conductivity, $\lambda$ the heat conductivity with subscripts *s* and *w* representing the formation solid and water respectively, $E_{mf} = h \cdot (T_f - T_m)$ the heat transport from fractures to matrix, $h$ the convective heat transport coefficient between matrix and fractures, and $Q_{hm}$ the heat source/sink term in matrix. Heat transport in the fractures of EGS is given as

$$b \rho_f C_f \frac{\partial T}{\partial t} + b \rho_f u_f C_f \nabla_T T - \nabla_T \cdot (b \lambda_f \nabla_T T) = E_{fm} + Q_{hf}$$

with $E_{fm} = h \cdot (T_m - T_f)$ the heat transport from matrix to fractures, and $Q_{hf}$ the heat source/sink term in fractures.

**Differential evolution.** Differential evolution (DE) is a stochastic metaheuristic algorithm for global optimization [59]. Before the evolution, $NP$ individuals $P = [\boldsymbol{x}_1, ..., \boldsymbol{x}_i, ..., \boldsymbol{x}_{NP}], i = \{1, ..., NP\}, \boldsymbol{x}_i \in \mathbb{R}^d$ are generated from the design space $[\boldsymbol{lb}, \boldsymbol{ub}]$, where $\boldsymbol{lb}$, $\boldsymbol{ub}$ are lower and upper bounds of search space. After population initialization, canonical DE works via 3 operators: mutation, crossover, and selection. The mutant vectors $[\boldsymbol{v}_1, ..., \boldsymbol{v}_i, ..., \boldsymbol{v}_{NP}], i = \{1, ..., NP\}, \boldsymbol{v}_i \in \mathbb{R}^d$ are generated with mutation operator:

$$\boldsymbol{v}_i = \boldsymbol{x}_{best} + Mu \times (\boldsymbol{x}_{r_1} - \boldsymbol{x}_{r_2})$$

with $r_1$, $r_2$ randomly generated integers from 1 to $NP$, $Mu$ mutation factor, and $\boldsymbol{x}_{best}$ the best solution in the population. Trial vectors are then generated using crossover operator

$$\boldsymbol{u}_i^j = \begin{cases} \boldsymbol{v}_i^j & if\ rand \leq CR\ or\ j = j_{rand} \\ \boldsymbol{x}_i^j & otherwise \end{cases}, j \in \{1, ..., d\}$$

with $\boldsymbol{u}_i^j$ the $j^{th}$ variable of the $i^{th}$ trial vector, $rand$ a random number from 0 to 1, and $CR$ crossover factor. The function value of trial vector $u$ is evaluated and new individual is selected via

$$x_i' = \begin{cases} u_i & if\ f(u_i) < f(x_i) \\ x_i & otherwise \end{cases}$$

with $x_i'$ the $i^{th}$ individual of the next generation.

**Probabilistic neural network.** In this study, we utilized a feed-forward neural network, identified as the probabilistic neural network (PNN) [60], as visually depicted in Fig. 2c, which involves an input layer, a pattern layer, a summation layer and an output layer. The respective numbers of neurons allocated to the input, pattern and output layers are denoted as $d$, $n$ and $l$, respectively. PNN exhibits superior computational efficiency with accurate prediction of target probability scores than ANN [61]. Unlike ANN, which heavily relies on activation functions such as sigmoid and rectified linear unit, and consequently finds it arduous to achieve admirable accuracy, PNN employs non-parametric functions in the form of the Gaussian kernel to determine pattern probability and enables classifying intricate patterns with promising precision.

## Data availability

All the datasets in the study are publicly available from the GitHub repository at https://github.com/JellyChen7/ALEMO.

## Code availability

Core codes developed in this study are publicly available from the GitHub repository at https://github.com/JellyChen7/ALEMO. The open-source evolutionary multi-objective optimization platform PlatEMO[62] is publicly available at https://github.com/BIMK/PlatEMO. The free open-source software packages MATLAB Reservoir Simulation Toolbox (MRST) for reservoir modelling and simulation is publicly available at https://www.sintef.no/projectweb/mrst/.

## Acknowledgement

This work was supported by the Research Grants Council of Hong Kong (CRF project C7082–22G), and the HKU Seed Fund for basic research. The computing resources provided by the High-Performance Computing Centre of Westlake University and kindly assisted by Zeqi Zheng are gratefully acknowledged.

## Competing interests

The authors declare no competing interests.

## Supplementary information

The supplementary information is attached.

Supplementary Information

# Machine Learning-Accelerated Multi-Objective Design of Fractured Geothermal Systems

Guodong Chen[1], Jiu Jimmy Jiao[1,*], Qiqi Liu[3], Zhongzheng Wang[2], Yaochu Jin[3]

[1] Department of Earth Sciences, The University of Hong Kong, Hong Kong, PR China.

[2] College of Engineering, Peking University, Beijing, PR China.

[3] College of Engineering, Westlake University, Hangzhou, PR China.

[*] Corresponding Author: Jiu Jimmy Jiao (jjiao@hku.hk).

Supplementary information includes:

CONTENTS

## Supplementary text

## Formulation and theorem

**Definition 1** (Pareto dominance): For any two solutions $\boldsymbol{x}_1, \boldsymbol{x}_2 \in \Omega$, $\boldsymbol{x}_1$ dominates $\boldsymbol{x}_2$ iff $f_i(\boldsymbol{x}_1) \leqslant f_i(\boldsymbol{x}_2)$ for $\forall i \in \{1, 2, ..., m\}$ and $f_j(\boldsymbol{x}_1) < f_j(\boldsymbol{x}_2)$ for $\exists j \in \{1, 2, ..., m\}$, referred to as $\boldsymbol{x}_1 \succ \boldsymbol{x}_2$.

**Definition 2** (Pareto optimal solutions and Pareto front): Any solution $\boldsymbol{x} \in \Omega$ is said to be Pareto optimal solution iff there is no other solution dominating it. Pareto optimal solutions in a set is known as Pareto set. The projection of the Pareto set into the objective space is known as Pareto front.

## Performance metrics

The Hypervolume (HV) indicator and Inverted Generational Distance (IGD) are commonly utilized performance metrics for accurately evaluating the convergence and diversity of the final population in a quantitative manner [1,2]. The IGD metric computes the average distance from each Pareto optimal solution to its nearest neighboring solution. Consider $\boldsymbol{P}^*$ as a collection of equidistant reference solutions on the Pareto front, and $\boldsymbol{Q}$ as a set of non-dominated solutions. Mathematically speaking, the IGD can be precisely defined as follows:

$$IGD(\boldsymbol{P}^*, \boldsymbol{Q}) = \frac{\sum_{\boldsymbol{p} \in \boldsymbol{P}^*} \mathrm{dis}(\boldsymbol{p}, \boldsymbol{Q})}{|\boldsymbol{P}^*|}$$

where $\mathrm{dis}(\boldsymbol{p}, \boldsymbol{Q})$ represents the minimum Euclidean distance between solution $\boldsymbol{p}$ and the set of solutions $\boldsymbol{Q}$ generated by MOEAs, and $|\boldsymbol{P}^*|$ represents the number of reference points. It is worth noting that a lower value of Inverted Generational Distance (IGD) signifies a more precise approximation to the true Pareto front. However, the calculation of IGD value needs the explicit distribution of Pareto front. When tackling real-world scientific and engineering problems, the explicit information pertaining to the distribution of the Pareto front is often unknown.

The Hypervolume (HV) indicator possesses the capacity to quantify the dominated volume of populations within the objective space without any prior knowledge about Pareto front distribution. The HV improvement is employed as the performance metric in multi-objective optimization problems. Assuming that the reference vector $\boldsymbol{z}_r = [z_{r1}, ..., z_{rm}]$ represents the reference point dominated by all Pareto optimal objective vectors, the HV is computed as:

$$HV(\boldsymbol{Q}, \boldsymbol{z}) = L([\boldsymbol{z}, \boldsymbol{x}_1] \cup ... \cup [\boldsymbol{z}, \boldsymbol{x}_k])$$

with $L$ the Lebesgue measure, and $\{\boldsymbol{x}_1, ..., \boldsymbol{x}_k\}$ the provided Pareto solutions in $\boldsymbol{Q}$.

## Preliminaries

**Radial basis function.** RBF have garnered widespread application in numerous scientific and industrial domains, owing to the remarkable capacity to approximate the landscape of objective functions using a small quantity of sample points.

As the method remains relatively insensitive to dimensionality and consistently shows promising approximation performance for non-linear, high-dimensional and correlated systems, RBF has been selected as the surrogate for this study. An RBF constitutes an interpolation method comprised of a weighted sum of basis functions:

$$\hat{f}(x) = \sum_{i=1}^{n} \omega_i \varphi(\|x - c_i\|)$$

where $\omega$ denotes the weight coefficient, $\varphi(\cdot)$ represents the basis function, $\|\cdot\|$ signifies the Euclidean norm, and $c$ corresponds to the centre points of the basis function. Explicitly, for a set of n training sample points $x = [x_1, ..., x_n]$ and corresponding objective function $y = [y_1, ..., y_n]$, $\omega$ can be ascertained as:

$$\omega = \varphi^{-1} y$$

the feature representations φ(***x***) are obtained by approximating a radial-basis function kernel, K(xi, xj) = exp(γ||xi−xj||$_2^2$), using Nystroem's method.

**Parameter settings.** The initial sample size for Latin hypercubic sampling is set at 100 for problems comprising fewer than 100 decision variables, while it is increased to 200 for those with 100 or more decision variables. The population size *NP* is set to 50. The parameters for polynomial mutation and binary crossover within the employed DE operator are set to 0.5 and 0.8, respectively. Polynomial mutation is applied with $\eta$ = 20 and $p_m$ = 1.

## Offline optimization versus online optimization

Offline optimization denotes the inability to actively query new samples during optimization, with offline surrogates exclusively constructed via one-shot samples. Conversely, online optimization signifies that additional sample points can be queried during optimization, with online surrogates updated employing various model management strategies [3]. To demonstrate the efficacy of online optimization, we compare the offline optimization with active learning enhanced online optimization (Supplementary Fig. S2). Efficient model management can yield considerably enhanced model quality, particularly in regions exhibiting high fitness values. The online surrogate is actively refined by querying new sample points within the reduced local promising area.

We visualize the landscape of 2-dimensional 2-objective DTLZ2 benchmark problem in Supplementary Fig. S2a with the corresponding Pareto front. After using design of experiment method for initial samples, we construct offline classifier surrogate with PNN and offline interpolation surrogate model with RBF in Supplementary Fig. S2b-c. The promising Class I area provided by the classifier is relatively large, since the samples in the promising area is sparse (Supplementary Fig. S2b). The RBF surrogate remains relatively high approximation error, although the main trend of the landscape is captured (Supplementary Fig. S2c). After querying new samples in the active learning process, the models are refined significantly. The promising Class I area provided by the updated classifier (Supplementary Fig. S2d) reduced significantly. The updated RBF surrogate (Supplementary Fig. S2e) also show superior approximation than original offline model.

## Discriminator versus regression-based surrogate

The PNN discriminator is essentially a classifier, aimed at identifying the region of current Pareto solutions. Its primary role is to classify solutions into the first rank of the Pareto front, selecting those with high uncertainty for real simulation evaluations. This infilling strategy focuses on exploring relatively promising areas characterized by high uncertainty. Meanwhile, the hypervolume-based attention subspace search strategy exploits high-fitness areas, leveraging the RBF surrogate model to accelerate convergence in the optimization process. These two infilling strategies are applied sequentially to effectively balance exploration (sampling from sparse and uncertain regions of the design space) and exploitation (sampling from high-fitness areas), thereby enhancing the overall optimization process. Traditional forward surrogate models primarily aim to approximate the objective function and often produce predictions that can lead to repetitive sampling, risking entrapment in local optima. Fig. S2 clearly illustrates the differences between the two machine learning models. The new training data is sequentially and strategically queried for efficiently discovering Pareto optimal solutions or close to Pareto optimal solutions with fewest number of samples.

## Supplementary Figures

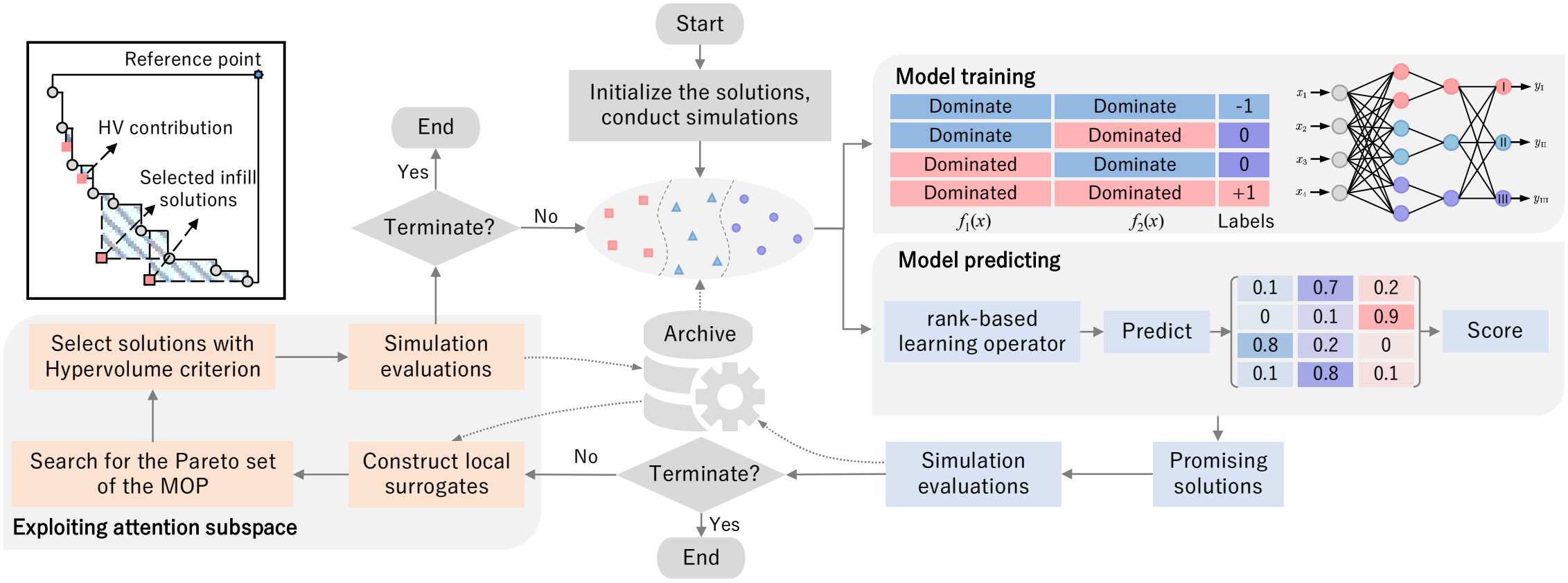

**Fig. S1. Schematic for the active learning enhanced evolutionary multi-objective optimization for computationally intensive geothermal systems.**

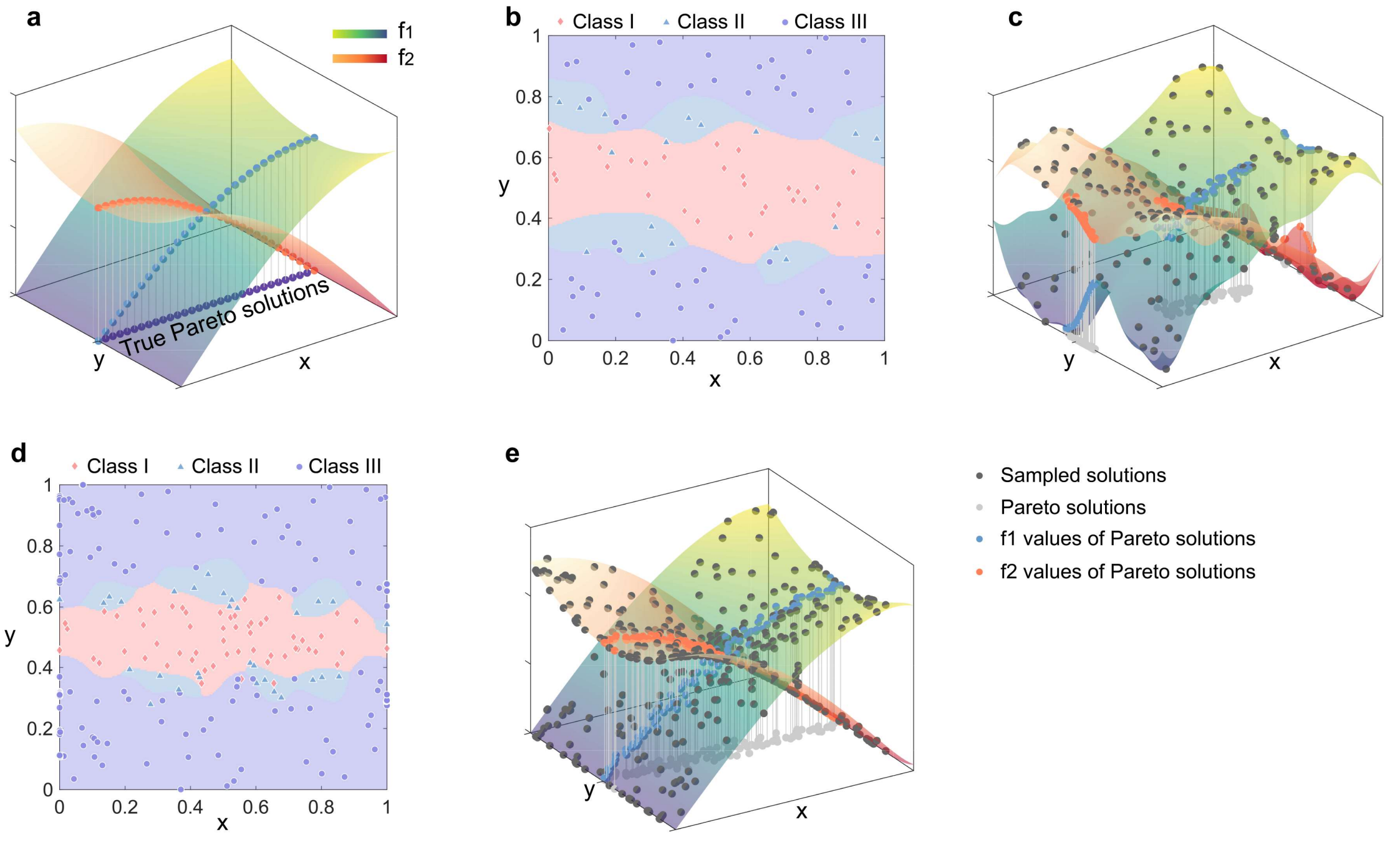

**Fig. S2. Comparison of offline optimization and active learning enhanced online optimization for benchmark problem. a**, True objective landscapes of $f_1$ and $f_2$ for DTLZ2 multi-objective benchmark problem. **b**, Offline classifier surrogate built with PNN using one-shot sample points. **c**, Offline surrogate modelling built with RBF method using one-shot sample points. **d**, Online classifier surrogate built with probabilistic neural network updated by active query of infill sample points. **e**, Online surrogate built with RBF method updated by active query of infill sample points.

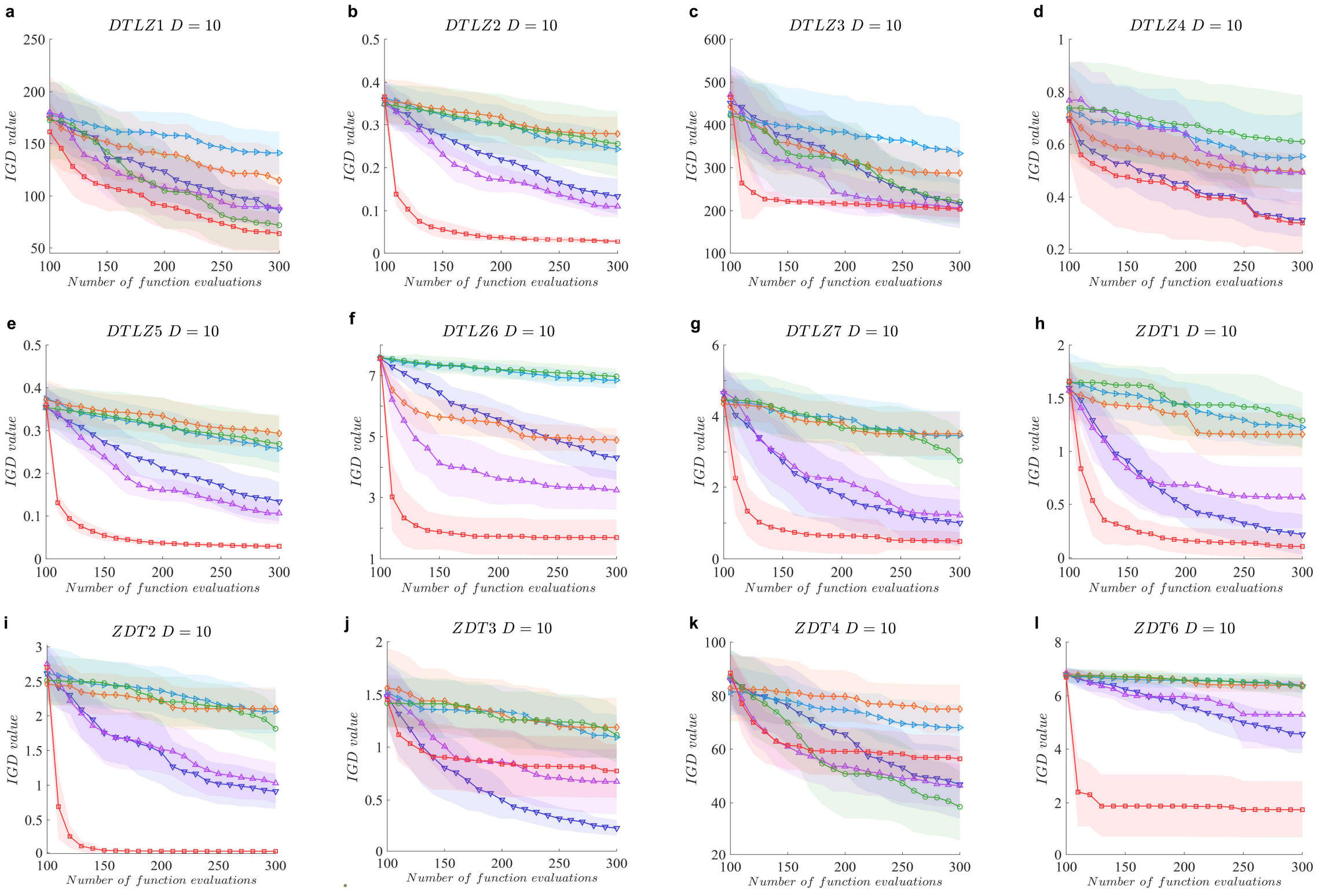

**Fig. S3.** Averaged convergence performance of data-driven evolutionary computation approaches on 10D benchmark problems with 20 independent trials presented as mean ± s.e.m.. **a-g**, Averaged convergence performance of data-driven evolutionary computation approaches on 10D DTLZ benchmark problems ($f_1$-$f_7$). **h-l**, Averaged convergence performance of data-driven evolutionary computation approaches on 10D ZDT benchmark problems ($f_8$-$f_{12}$).

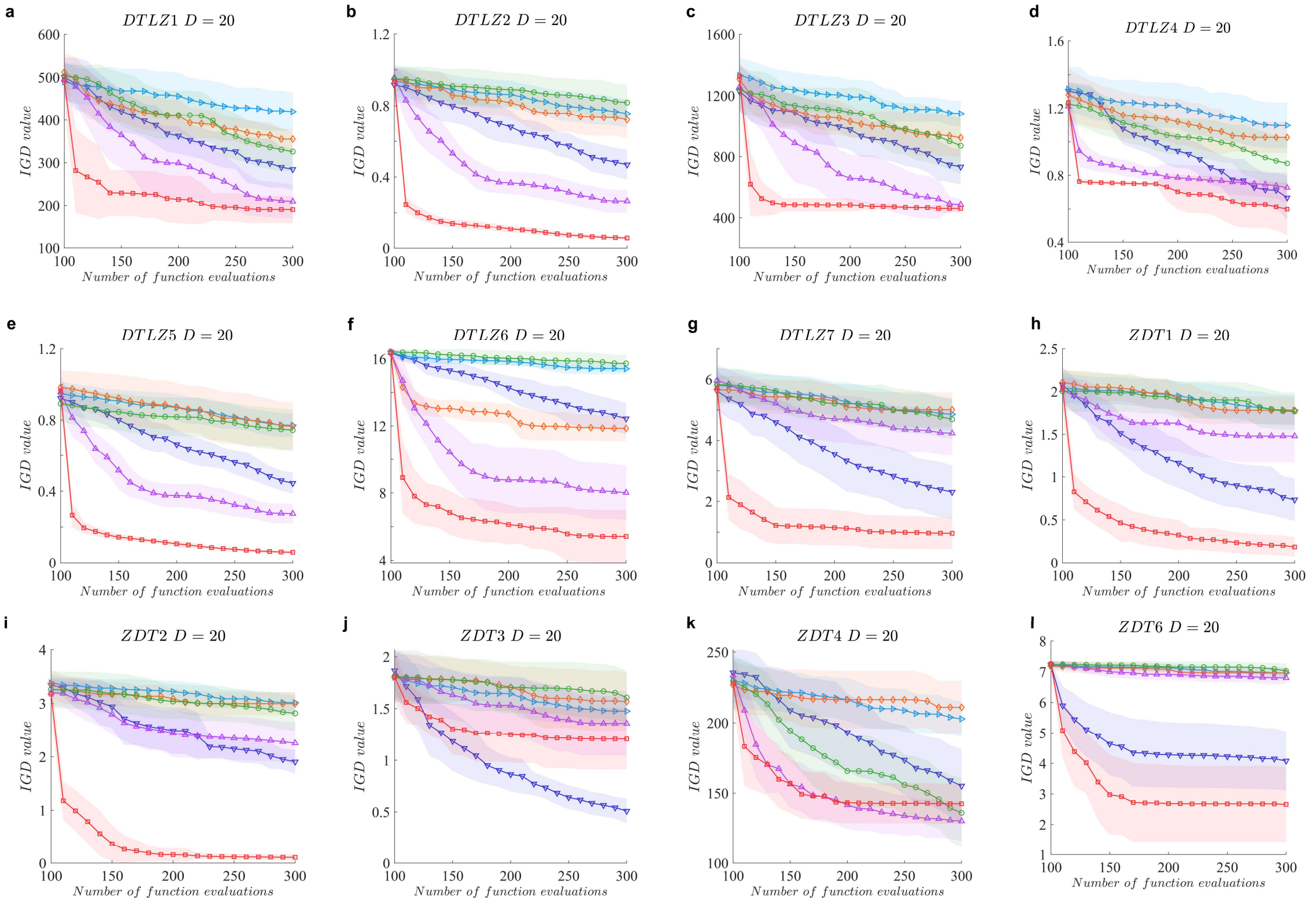

**Fig. S4.** Averaged convergence performance of data-driven evolutionary computation approaches on 20D benchmark problems with 20 independent trials presented as mean ± s.e.m.. **a-g**, Averaged convergence performance of data-driven evolutionary computation approaches on 20D DTLZ benchmark problems ($f_1$-$f_7$). **h-l**, Averaged convergence performance of data-driven evolutionary computation approaches on 20D ZDT benchmark problems ($f_8$-$f_{12}$).

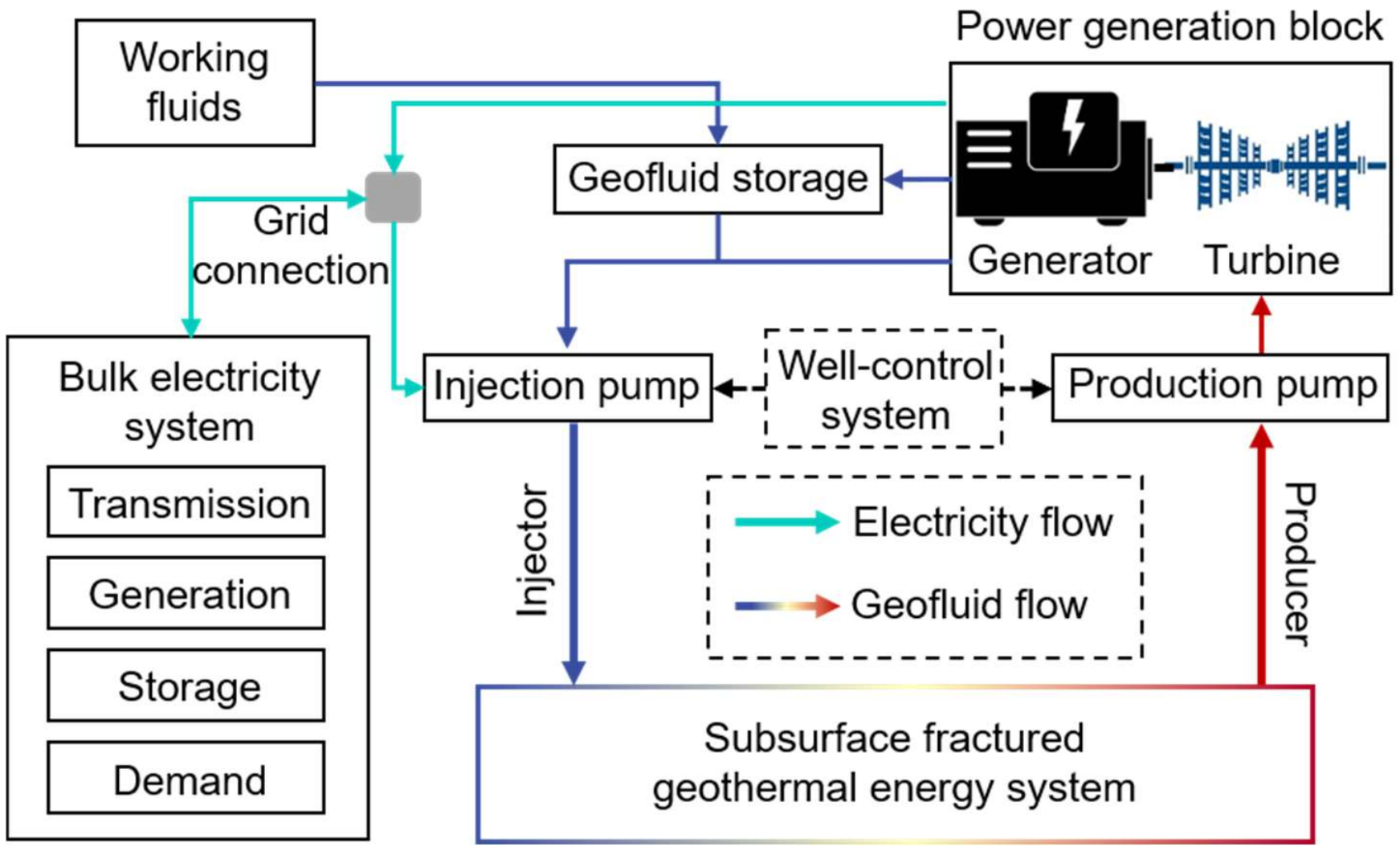

**Fig. S5. Schematic diagram of the geothermal power generation systems.** The connection between the geothermal power plant and the bulk electricity system. The aim of ALEMO in the well-control system is to provide optimal well-control schemes to maximize the profits according to the preference of decision-makers.

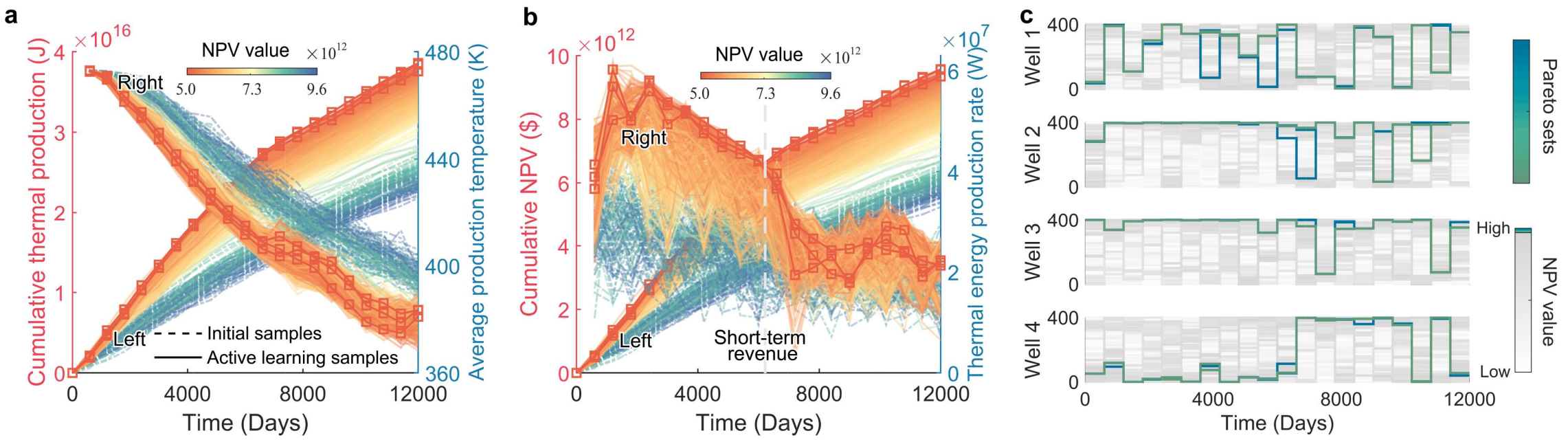

**Fig. S6. Results of ALEMO and evolution of thermal dynamics of the fractured geothermal reservoir. a**, Simulation results of the cumulative thermal energy production and average production temperature of the enhanced geothermal energy system for the sampling history of ALEMO. **b**, Simulation results of the cumulative net present value and thermal energy production rate for the sampling history of ALEMO. **c**, Well control trajectories and corresponding net present value of sampling history of ALEMO.

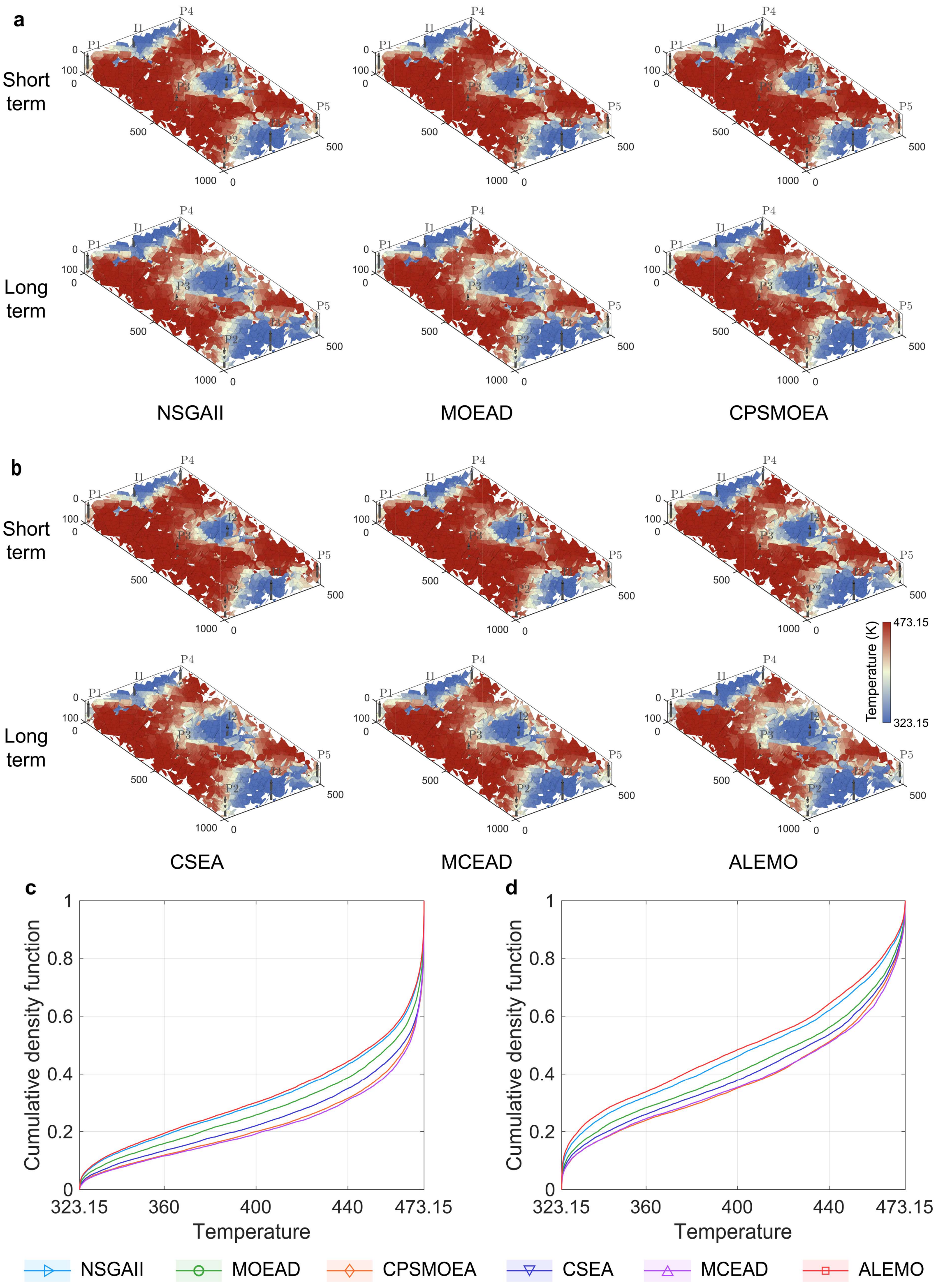

**Fig. S7. Simulation results of ALEMO and compared baseline methods. a**, Simulation results of the cumulative thermal energy production and average production temperature of the enhanced geothermal energy system for the sampling history of ALEMO.

**evolution of thermal dynamics of the fractured geothermal reservoir. a**, Simulation results of the cumulative thermal energy production and average production temperature of the enhanced geothermal energy system for the sampling history of ALEMO. **b**, Simulation results of the cumulative net present value and thermal energy production rate for the sampling history of ALEMO. **c**, Well control trajectories and corresponding net present value of sampling history of ALEMO.

## Supplementary Tables

Supplementary Table S-I Characteristics of the DTLZ multi-objective benchmark problems

| Problem | Function Expression | M | D | Separability | Modality | Bias | Geometry |
|---|---|---|---|---|---|---|---|
| DTLZ1 ($f_1$) | $\min\ f_1(\boldsymbol{x}) = \frac{1}{2}x_1 x_2(1+g(\boldsymbol{x}_M))$<br>$\min\ f_2(\boldsymbol{x}) = \frac{1}{2}x_1(1-x_2)(1+g(\boldsymbol{x}_M))$<br>......<br>$\min\ f_{M-1}(\boldsymbol{x}) = \frac{1}{2}x_1(1-x_2)(1+g(\boldsymbol{x}_M))$<br>$\min\ f_M(\boldsymbol{x}) = \frac{1}{2}(1-x_1)(1+g(\boldsymbol{x}_M))$<br>$with\ g(\boldsymbol{x}_M) = 100\left[\lvert\boldsymbol{x}_M\rvert + \sum_{x_i \in \boldsymbol{x}_M}(x_i-0.5)^2 - \cos(20\pi(x_i-0.5))\right],\ \boldsymbol{x}_M = [x_M,...,x_L$<br>$s.t.,\ 0 \le x_i \le 1, i = 1,2,...,D$ | 2 | 10, 20, 30 | Separable | M | No | Linear |
| DTLZ2 ($f_2$) | $\min\ f_1(\boldsymbol{x}) = (1+g(\boldsymbol{x}_M))\cos(\frac{\pi x_1}{2})...\cos(\frac{\pi x_{M-2}}{2})\cos(\frac{\pi x_{M-1}}{2})$<br>$\min\ f_2(\boldsymbol{x}) = (1+g(\boldsymbol{x}_M))\cos(\frac{\pi x_1}{2})...\cos(\frac{\pi x_{M-2}}{2})\sin(\frac{\pi x_{M-1}}{2})$<br>......<br>$\min\ f_M(\boldsymbol{x}) = (1+g(\boldsymbol{x}_M))\sin(\frac{\pi x_1}{2})$<br>$with\ g(\boldsymbol{x}_M) = \sum_{x_i \in \boldsymbol{x}_M}(x_i-0.5)^2,\ \boldsymbol{x}_M = [\boldsymbol{x}_M,...,\boldsymbol{x}_D]$<br>$s.t.,\ 0 \le x_i \le 1, i = 1,2,...,D$ | 2 | 10, 20, 30 | Non-separable | U | No | Concave |
| DTLZ3 ($f_3$) | $\min\ f_1(\boldsymbol{x}) = (1+g(\boldsymbol{x}_M))\cos(\frac{\pi x_1}{2})...\cos(\frac{\pi x_{M-2}}{2})\cos(\frac{\pi x_{M-1}}{2})$<br>$\min\ f_2(\boldsymbol{x}) = (1+g(\boldsymbol{x}_M))\cos(\frac{\pi x_1}{2})...\cos(\frac{\pi x_{M-2}}{2})\sin(\frac{\pi x_{M-1}}{2})$<br>......<br>$\min\ f_M(\boldsymbol{x}) = (1+g(\boldsymbol{x}_M))\sin(\frac{\pi x_1}{2})$<br>$with\ g(\boldsymbol{x}_M) = 100\left[\lvert\boldsymbol{x}_M\rvert + \sum_{x_i \in \boldsymbol{x}_M}(x_i-0.5)^2 - \cos(20\pi(x_i-0.5))\right],\ \boldsymbol{x}_M = [x_M,...,x_D]$<br>$s.t.,\ 0 \le x_i \le 1, i = 1,2,...,D$ | 2 | 10, 20, 30 | Separable | M | No | Concave |
| DTLZ4 ($f_4$) | $\min\ f_1(\boldsymbol{x})=(1+g(\boldsymbol{x}_M))\cos(\frac{\pi x_1^{\alpha}}{2})\ldots\cos(\frac{\pi x_{M-2}^{\alpha}}{2})\cos(\frac{\pi x_{M-1}^{\alpha}}{2})$<br>$\min\ f_2(\boldsymbol{x})=(1+g(\boldsymbol{x}_M))\cos(\frac{\pi x_1^{\alpha}}{2})\ldots\cos(\frac{\pi x_{M-2}^{\alpha}}{2})\sin(\frac{\pi x_{M-1}^{\alpha}}{2})$<br>......<br>$\min\ f_M(\boldsymbol{x})=(1+g(\boldsymbol{x}_M))\sin(\frac{\pi x_1^{\alpha}}{2})$<br>*with* $g(\boldsymbol{x}_M)=\sum_{x_i\in\boldsymbol{x}_M}(x_i-0.5)^2,\ \boldsymbol{x}_M=[x_M,\ldots,x_D]$<br>*s.t.*, $0\le x_i\le 1, i=1,2,\ldots,D$ | 2 | 10, 20, 30 | Separable | U | Yes | Concave, many-to-one |
| DTLZ5 ($f_5$) | $\min\ f_1(\boldsymbol{x})=(1+g(\boldsymbol{x}_M))\cos(\frac{\pi\theta_1}{2})\ldots\cos(\frac{\pi\theta_{M-2}}{2})\cos(\frac{\pi\theta_{M-1}}{2})$<br>$\min\ f_2(\boldsymbol{x})=(1+g(\boldsymbol{x}_M))\cos(\frac{\pi\theta_1}{2})\ldots\cos(\frac{\pi\theta_{M-2}}{2})\sin(\frac{\pi\theta_{M-1}}{2})$<br>......<br>$\min\ f_M(\boldsymbol{x})=(1+g(\boldsymbol{x}_M))\sin(\frac{\pi\theta_1}{2})$<br>*with* $\theta_i=\frac{\pi}{4(1+g(\boldsymbol{x}_M))}(1+2g(\boldsymbol{x}_M)x_i),\ i=2,3,\ldots,(M-1)$<br>$g(\boldsymbol{x}_M)=\sum_{x_i\in\boldsymbol{x}_M}(x_i-0.5)^2,\ \boldsymbol{x}_M=[x_M,\ldots,x_D]$<br>*s.t.*, $0\le x_i\le 1, i=1,2,\ldots,D$ | 2 | 10, 20, 30 | Separable | U | Yes | Concave |
| DTLZ6 ($f_6$) | $\min\ f_1(\boldsymbol{x})=(1+g(\boldsymbol{x}_M))\cos(\frac{\pi\theta_1}{2})\ldots\cos(\frac{\pi\theta_{M-2}}{2})\cos(\frac{\pi\theta_{M-1}}{2})$<br>$\min\ f_2(\boldsymbol{x})=(1+g(\boldsymbol{x}_M))\cos(\frac{\pi\theta_1}{2})\ldots\cos(\frac{\pi\theta_{M-2}}{2})\sin(\frac{\pi\theta_{M-1}}{2})$<br>......<br>$\min\ f_M(\boldsymbol{x})=(1+g(\boldsymbol{x}_M))\sin(\frac{\pi\theta_1}{2})$<br>*with* $\theta_i=\frac{\pi}{4(1+g(\boldsymbol{x}_M))}(1+2g(\boldsymbol{x}_M)x_i),\ i=2,3,\ldots,(M-1)$<br>$g(\boldsymbol{x}_M)=\sum_{x_i\in\boldsymbol{x}_M}x_i^{0.1},\ \boldsymbol{x}_M=[x_M,\ldots,x_D]$<br>*s.t.*, $0\le x_i\le 1, i=1,2,\ldots,D$ | 2 | 10, 20, 30 | Non-separable | U | Yes | Concave |
| DTLZ7 ($f_7$) | $\min\ f_1(\boldsymbol{x})=x_1$<br>$\min\ f_2(\boldsymbol{x})=x_2$<br>......<br>$\min\ f_M(\boldsymbol{x})=(1+g(\boldsymbol{x}_M))h(f_1,f_2,\ldots,f_{M-1},g)$<br>*with* $g(\boldsymbol{x}_M)=1+\frac{9}{\lvert\boldsymbol{x}_M\rvert}\sum_{x_i\in\boldsymbol{x}_M}x_i,\ \boldsymbol{x}_M=[x_M,\ldots,x_D]$<br>$h(f_1,f_2,\ldots,f_{M-1},g)=M-\sum_{i=1}^{M-1}\left[\frac{f_i}{1+g}(1+\sin(3\pi f_i))\right]$<br>*s.t.*, $0\le x_i\le 1, i=1,2,\ldots,D$ | 2 | 10, 20, 30 | Non-separable | M | Yes | Disconnected |

Supplementary Table S-II Characteristics of the ZDT multi-objective benchmark problems

| Problem | Function Expression | M | D | Separability | Modality | Bias | Geometry |
|---|---|---|---|---|---|---|---|
| ZDT1 ($f_8$) | $\min f_1(\boldsymbol{x}) = x_1$<br>$\min f_2(\boldsymbol{x}) = (1+\frac{9}{D-1}\sum_{i=2}^{D} x_i)(1-\sqrt{\frac{x_1}{1+\frac{9}{D-1}\sum_{i=2}^{D} x_i}})$<br>$s.t.,\ 0 \le x_i \le 1, i = 1,2,...,D$ | 2 | 10, 20, 30 | Separable | U | No | Convex |
| ZDT2 ($f_9$) | $\min f_1(\boldsymbol{x}) = x_1$<br>$\min f_2(\boldsymbol{x}) = (1+\frac{9}{D-1}\sum_{i=2}^{D} x_i)(1-\frac{x_1}{1+\frac{9}{D-1}\sum_{i=2}^{D} x_i})^2$<br>$s.t.,\ 0 \le x_i \le 1, i = 1,2,...,D$ | 2 | 10, 20, 30 | Separable | U | No | Concave |
| ZDT3 ($f_{10}$) | $\min f_1(\boldsymbol{x}) = x_1$<br>$\min f_2(\boldsymbol{x}) = (1+\frac{9}{D-1}\sum_{i=2}^{D} x_i)[1-\sqrt{\frac{x_1}{1+\frac{9}{D-1}\sum_{i=2}^{D} x_i}}-\frac{x_1}{1+\frac{9}{D-1}\sum_{i=2}^{D} x_i}\sin(10\pi x_1)]$<br>$s.t.,\ 0 \le x_i \le 1, i = 1,2,...,D$ | 2 | 10, 20, 30 | Separable | M | No | Convex, disconnected |
| ZDT4 ($f_{11}$) | $\min f_1(\boldsymbol{x}) = x_1$<br>$\min f_2(\boldsymbol{x}) = [1+10(D-1)+\sum_{i=2}^{D}(x_i^2-10\cos(4\pi x_i))][1-\sqrt{\frac{x_1}{1+10(D-1)+\sum_{i=2}^{D}(x_i^2-10\cos(4\pi x_i))}}$<br>$s.t.,\ 0 \le x_i \le 1, i = 1,2,...,D$ | 2 | 10, 20, 30 | Separable | M | No | Convex, multi-modal |
| ZDT6 ($f_{12}$) | $\min f_1(\boldsymbol{x}) = 1-\exp(-4x_1)\sin^6(6\pi x_1)$<br>$\min f_2(\boldsymbol{x}) = [1+9[(\sum_{i=2}^{D} x_i)/9]^{0.25}][1-(\frac{x_1}{1+9[(\sum_{i=2}^{D} x_i)/9]^{0.25}})^2]$<br>$s.t.,\ 0 \le x_i \le 1, i = 1,2,...,D$ | 2 | 10, 20, 30 | Separable | M | Yes | Convex, multi-modal, many-to-one |

Supplementary Table S-III Parameter configurations of the fractured geothermal reservoir

| Parameter | Value |
|---|---|
| The initial temperature | 200 ℃ |
| The initial pressure | 30 MPa |
| Temperature of injection fluids | 20 ℃ |
| BHP of producers | 30MPa |
| Flow rate of injectors | [0, 400] $\times 10^{-3} m^3/s$ |
| Depth of the model | 2500m |
| Rock matrix porosity | 0.01 |
| Fracture porosity | 0.1 |
| Rock matrix permeability | $5\times 10^{-15} m^2$ |
| Permeability of fractures | $10^{-7}$ $m^2$ |
| Reservoir thickness | 40 m |
| Model matrix thermal conductivity | 2 W/(m*K) |
| Fluid thermal conductivity | 0.698 W/(m*℃) |
| Matrix heat capacity | 850 J/(kg ℃) |
| Fluid heat capacity | 4200 J/(kg ℃) |